\documentclass[lettersize,journal]{IEEEtran}
\usepackage{amsmath,amsfonts}
\usepackage{algorithmic}
\usepackage{algorithm}
\usepackage{array}
\usepackage[caption=false,font=normalsize,labelfont=sf,textfont=sf]{subfig}
\usepackage{textcomp}
\usepackage{stfloats}
\usepackage{url}
\usepackage{verbatim}
\usepackage{graphicx}
\usepackage{cite}
\usepackage{afterpage}
\usepackage{graphicx}
\usepackage{amsmath}
\usepackage{amssymb}
\usepackage{booktabs}
\usepackage{svg}
\usepackage{tabularx}
\usepackage{multirow}
\usepackage{ragged2e}
\usepackage{lipsum}
\usepackage[numbers,sort,compress]{natbib}

\begin{document}

% \linenumbers
\title{Interpretable Concept-based Deep Learning Framework for Multimodal Human Behavior Modeling}

% \author{Xinyu Li,~\IEEEmembership{Student Member,~IEEE,} Marwa Mahmoud,~\IEEEmembership{Member,~IEEE}
\author{\parbox{16cm}{\centering
   {\large Xinyu Li and Marwa Mahmoud }\\
   {\normalsize
   School of Computing Science, University of Glasgow, United Kingdom\\}}
}
        % <-this % stops a space



\maketitle

\begin{abstract}

In the contemporary era of intelligent connectivity, Affective Computing (AC), which enables systems to recognize, interpret, and respond to human behavior states, has become an integrated part of many AI systems. As one of the most critical components of responsible AI and trustworthiness in all human-centered systems, explainability has been a major concern in AC. Particularly, the recently released EU General Data Protection Regulation requires any high-risk AI systems to be sufficiently interpretable, including biometric-based systems and emotion recognition systems widely used in the affective computing field. Existing explainable methods often compromise between interpretability and performance. Most of them focus only on highlighting key network parameters without offering meaningful, domain-specific explanations to the stakeholders. Additionally, they also face challenges in effectively co-learning and explaining insights from multimodal data sources. To address these limitations, we propose a novel and generalizable framework, namely the Attention-Guided Concept Model (AGCM), which provides learnable conceptual explanations by identifying \textit{what} concepts that lead to the predictions and \textit{where} they are observed. AGCM is extendable to any spatial and temporal signals through multimodal concept alignment and co-learning, empowering stakeholders with deeper insights into the model's decision-making process. We validate the efficiency of AGCM on well-established Facial Expression Recognition benchmark datasets while also demonstrating its generalizability on more complex real-world human behavior understanding applications. 
We believe that AGCM’s flexibility and extensibility lay a solid foundation for developing future interpretable and trustworthy models in downstream affective computing applications, including in mental health, psychiatry, education, automotive, and security, offering both competitive performance and domain-specific explanations.

\end{abstract}


\section{Introduction}
    % \IEEEPARstart{E}{xplainability}
    Affective Computing (AC) aims to develop models and systems that recognize, interpret, and respond to human behavior states. As a human-centered design, explainability and transparency have become critical concerns in AC applications \cite{cortinas2023toward}. The EU AI Act \cite{hupont2022landscape} and the newly proposed General Data Protection Regulation (GDPR) in 2024 \cite{GDPR2024} mandates that high-risk AI systems, including biometric-based systems and emotion recognition systems widely used in the affective computing field, must be sufficiently transparent to allow stakeholders from cross-disciplinary area to comprehend the decision-making process of the framework. Enhancing explainability in AC models not only offers extra insights into AI predictions but also ensures fair, trustworthy, and accountable outcomes in sensitive applications like education, healthcare, and security systems. \cite{yu2024bridging, kumar2023opacity}.

    \begin{figure}[t]
    \centering
    \includegraphics[width=0.99\columnwidth]{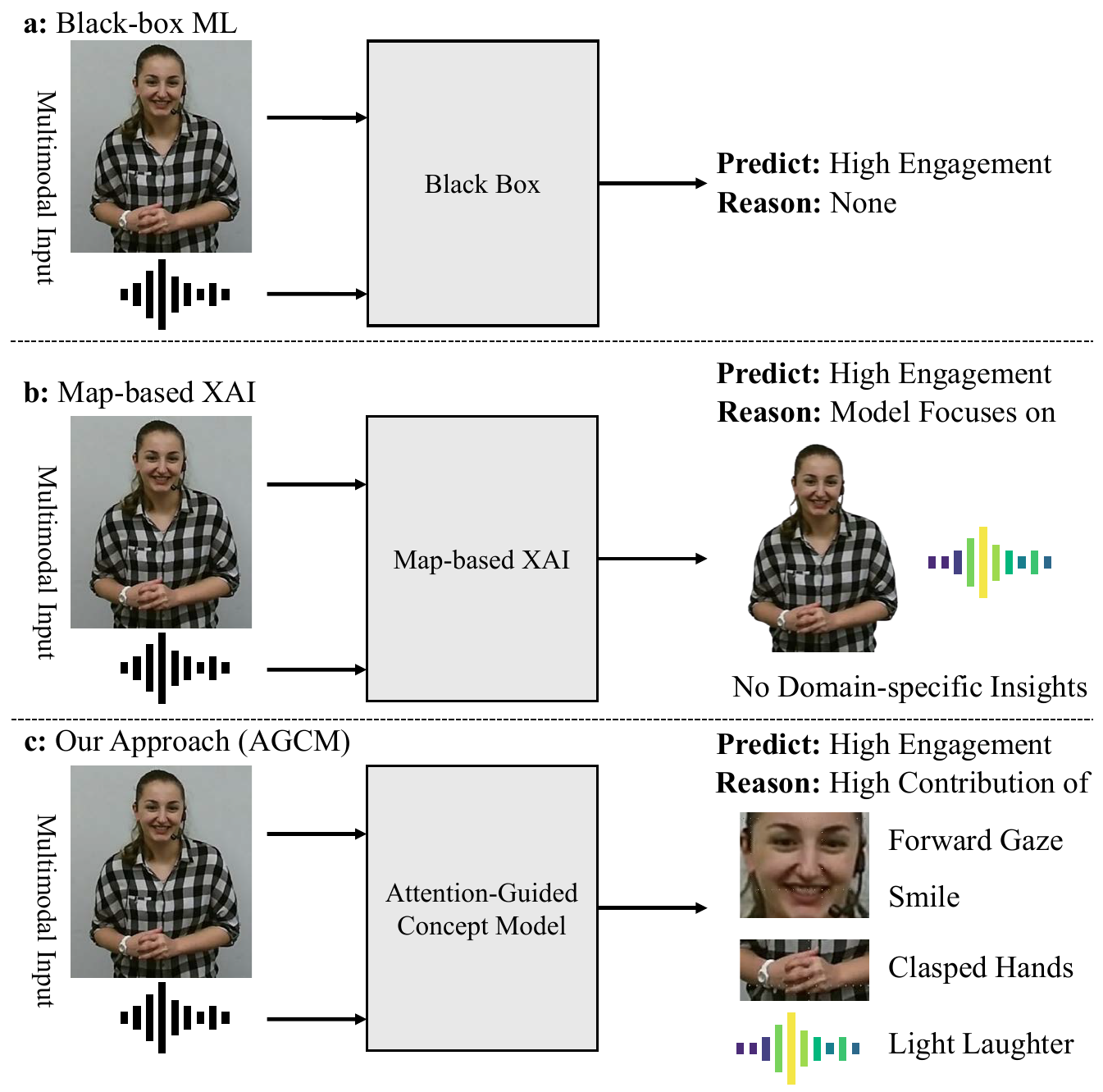}
       \caption{Difference between the black-box models, current eXplainable AI (XAI), and our proposed model. (a): Black-box ML models offer no extra insight into the model prediction. (b): Map-based XAI approaches offer explanations by identifying important regions that lead to the prediction, but without any domain-specific knowledge that validates the decision-making process. (c): Our proposed framework explicitly localizes domain-specific indicators, learns their contributions during training, and incorporates multimodal concepts, thereby making predictions based on these intermediate attributes in an inherently interpretable manner.}
    \label{fig_intro}
    \end{figure}

    There is an increasing interest in developing interpretable or eXplainable Artificial Intelligence (XAI) to improve model transparency in AC. As shown in Fig. \ref{fig_intro} (b), approaches such as post-hoc explanations \cite{ribeiro2016should,heimerl2020unraveling,malik2021towards} and map-based methods \cite{gao2021ts,gund2021interpretable,belharbi2024guided} have emerged to address this need. However, these techniques primarily focus on identifying important regions or parameters within deep neural networks, rather than providing an explicit, causal explanation for the predictions. This limitation is especially pronounced in AC, where opposing facial Action Units, like AU12 (Lip Corner Puller) associated with positive emotion and AU15 (Lip Corner Depressor) linked to negative emotion, can occur in the same facial region. Meanwhile, the alignment and co-learning from multimodal sources pose even greater challenges for these approaches due to the inherently different properties of multimodal knowledge. Therefore, they often face a trade-off between performance and interpretability, which, in high-risk XAI, may undermine the system's trustworthiness \cite{rudin2019stop}.

    Consider a common question: How would a human expert explain their prediction of an individual with highly conversational engagement? They would likely point to the activation of specific facial muscles, such as the zygomatic major, indicating engaged smiles, a strong positive indicator of engagement. Meanwhile, the forward gaze direction, proper gesture, body language, and audio indicators can also be used to recognize engagement. Thus, a good explanation from an AC model should address two key aspects: \textit{what} indicators or concepts (e.g., facial muscle activations) contribute to the prediction, and \textit{where} these concepts are observed. Furthermore, the importance of multimodal learning is self-evident in real-world AC applications \cite{abaeikoupaei2020multi, yoon2022d, cafaro2017noxi}. Training and interpreting AC models with multimodal alignment and co-learning is another key challenge in affective XAI \cite{baltruvsaitis2018multimodal}.

    As shown in Fig. \ref{fig_intro} (c), in this paper, we propose an interpretable concept-based framework: the Attention-Guided Concept Model (AGCM), which localizes and learns the key indicators during training and then makes the final prediction according to the contribution of these intermediate concepts. This framework incorporates spatial concept information and multimodal concept fusion within a powerful attention-based architecture, combining the advantage of both domain-specific explanation and state-of-the-art performance. In summary, the main contributions of this paper are as follows:

    \begin{enumerate}
    \item We propose a concept-based interpretable framework for AC applications, namely the Attention-Guided Concept Model (AGCM), which provides both learnable multimodal conceptual explanations and spatial visual concept localization, quantifying the contribution of individual concepts to the predicted affective label.
    \item To address the challenge of multimodal concept alignment and co-learning, AGCM introduces an extendable sequential multimodal concept fusion, which can be easily expanded to any spatial-temporal signal. This approach accounts for temporal and contextual information between input modalities, demonstrating the adaptability to other discrete or continuous signals.
    \item We qualitatively and quantitatively evaluate the proposed framework on three large-scale FER datasets: RAF-DB, AffectNet, and Aff-Wild2, demonstrating that AGCM outperforms previous interpretable models and achieves competitive performance compared to state-of-the-art black-box models. Moreover, the experiment shows that AGCM offers a human-interpretable explanation grounded in domain-specific knowledge.
    \item To demonstrate the generalizability of AGCM on complex real-world AC applications, we conduct extensive experiments on the human-human interaction dataset, validating its ability to provide explainable and accurate prediction in downstream AC applications. We provide a video demonstration in the supplementary material to offer additional insights into the prediction process and its explainability.
    \end{enumerate}

\section{Related Work}

In this section, we examine two primary machine learning approaches commonly used in affective computing: feature-based models and end-to-end models. We then discuss recent advancements in explainable affective computing, emphasizing their contribution and limitation to model transparency and interpretability.

\subsection{Feature and End-to-end Models in Affective Computing}
    Discriminative AC focuses on mapping human-centered data to emotion-related labels, employing two primary approaches: feature-based models and end-to-end models.
    
    Feature-based models \cite{tsalera2022feature, avola2022affective} rely on manually extracted features derived from raw data, which are then used to train machine learning models to establish the relationship between features and labels. The strength of this approach lies in the interpretability of the features, which are often human-understandable and can provide valuable behavioral insights \cite{bento2022comparing}. Additionally, feature-based models typically operate on structured, tabular data, offering a computationally efficient solution \cite{bisogni2023emotion}. However, the reliance on handcrafted features may omit potentially important information embedded in the raw data, causing inevitable information loss \cite{zhao2019affective, cortinas2023toward}. Furthermore, decoupling feature extraction from model training may introduce limitations, such as overfitting, particularly due to the structured nature of the input data \cite{bengio2013representation}.
    
    End-to-end models \cite{li2020deep}, on the other hand, learn directly from raw data, eliminating the need for manual feature engineering. Fully leveraging the representational power of deep neural networks, these models are particularly effective when trained on large datasets. However, their strength is also their weakness: the opacity of their learned representations often leads to what is referred to as the ``black-box'' problem, making these models difficult to interpret as they lack human-understandable intermediate representations \cite{zhao2019affective}. 
    This challenge persists in multi-task learning, where models are designed to predict multiple task labels simultaneously, such as emotion and AUs. Despite their multi-task design, emotion and AU predictions are learned independently, leaving the model as a black box, where the predicted AUs cannot explain the predicted emotions.

\begin{figure}[t]
    \centering
    \includegraphics[width=0.7\columnwidth]{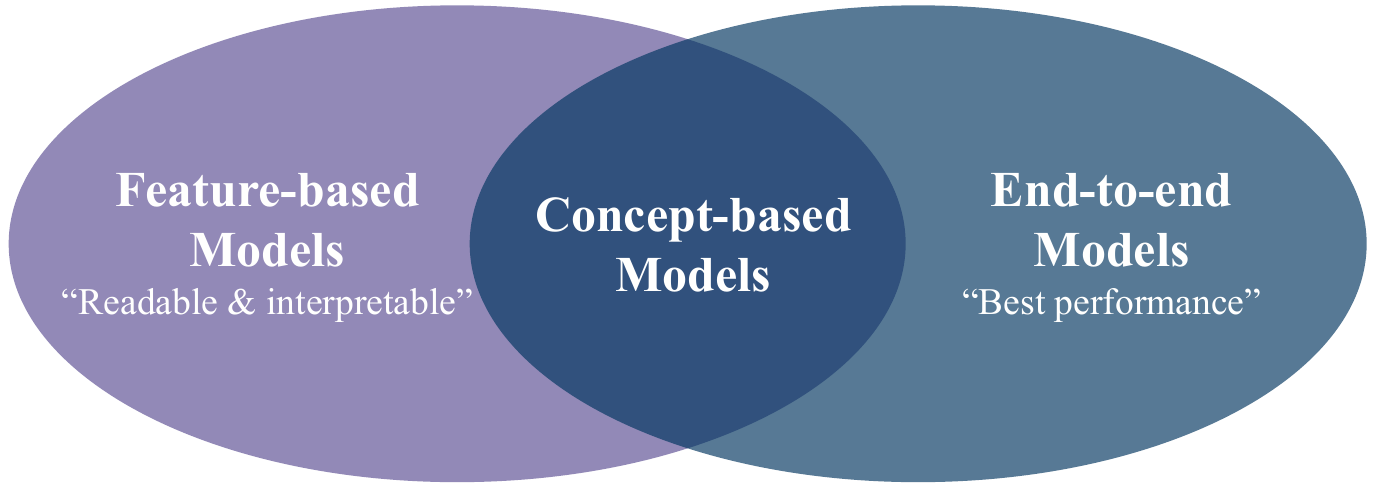}
       \caption{Feature-based approaches offer inherent interpretability and are easily understood by humans, while end-to-end models deliver state-of-the-art learning capabilities. This work seeks to integrate the strengths of both methods through a concept-based framework, which achieves a balance between high explainability and robust performance. Unlike traditional features, concepts are not static values. They serve as the neural embeddings of features that are trainable within the ML framework, spontaneously quantifying the contribution of individual concepts to the task label. }
    \label{fig_ven}
    \end{figure}

    As shown in Fig. \ref{fig_ven}, in this work, we propose a hybrid approach, integrating the strengths of the well-understandable feature-based model and the state-of-the-art black-box models through concept-based learning, where each concept serves as an embedded neural representation of the feature. This approach retains the interpretability inherent in feature-based models while harnessing the robust learning capabilities of end-to-end neural networks. 

\subsection{XAI in Affective Computing} \label{xai_in_ac}
    Recent efforts to enhance the explainability of affective computing models have largely relied on post-hoc, map-based visualizations, and concept-based learning.

    Post-hoc approaches \cite{ribeiro2016should,heimerl2020unraveling,malik2021towards, Wu2019Enhancing} retrospectively analyze the parameter importance of pre-trained black-box models after deployment. These methods attempt to explain the model by manipulating parameters in specific parts of the network to check their impact on the final prediction. Map-based approaches \cite{gund2021interpretable,belharbi2024guided} are another common method used to interpret black-box models, typically highlighting the regions where the model focuses its attention. However, both of these approaches primarily focus on the importance scores within the neural network, without offering additional, domain-relevant information for experts. This limitation is particularly evident in AC, where conflicting indicators, such as AU12 (Lip Corner Puller) signaling positive emotion and AU15 (Lip Corner Depressor) indicating negative emotion, may appear in the same facial region. Therefore, simply presenting the weight importance or model attention provides little insight for domain experts like psychologists to understand the AI decision-making process. Furthermore, the distinct properties of multimodal data make incorporating multimodal alignment and co-learning in post-hoc or map-based XAI methods even more challenging, taking the risk of losing either accuracy or interpretability.

    Recent attempts on concept-based models \cite{xinyuFG24, zarlenga2022concept} try to encapsulate specific, human-understandable features through concept embeddings $C$ that are learned in a fully supervised manner. These models learn the mapping $X \rightarrow C \rightarrow Z$, where $x \in X$ represents the raw image pixels and $z \in Z$ represents the task labels. Specifically, a concept generator $G$ generates concept embeddings, denoted as $\hat{c} = G(x)$, with $\hat{c} \in C$ representing the learned concepts within a bottleneck layer $C$. Subsequently, a facial expression predictor $y$ maps the concept embeddings to task labels $\hat{z} \in Z$, where $\hat{z} = y(\hat{c})$. While concept-based models offer a more interpretable framework than map-based approaches, ongoing research is focused on integrating this explainable architecture with multimodal learning and performance-optimized strategies \cite{xinyuFG24}. Moreover, a key challenge lies in integrating spatial explanations, which reveal \textit{where} the model is focusing, with concept-based explanations, which clarify \textit{what} contributes to the prediction. Achieving this synergy is essential for enhancing both the interpretability and practical utility of models in high-stakes applications.

    Table \ref{tab_previous_work} compares the proposed concept-based framework with previous feature-based, map-based, and black-box FER models in terms of explainability and performance. The proposed framework provides learnable domain-specific insights into the decision-making process for stakeholders while retaining map-based explanations that illustrate the model's areas of attention. A two-stage learning architecture with multimodal concept fusion is introduced, effectively addressing the alignment and co-learning challenges in multimodal interpretable AC. Furthermore, it achieves state-of-the-art performance through deep end-to-end training, successfully balancing the trade-off between interpretability and performance in high-stakes AC applications.

    \begin{table}[t]
    \centering
    \caption{Comparison of our work with previous works on FER in terms of explainability and performance, including feature-based approach, map-based approach, and deep end-to-end approach.}
    \begin{tabular}{ccccc}
    \toprule
                          & Ours & Feature & Map & Black-box \\ \midrule
    Feature-based Insight & +    & +       &     &     \\
    Map-based Explanation & +    &         & +   &     \\
    End-to-end Training   & +    &         & +   & +   \\
    Learnable Explanation & +    &         &     &     \\
    Multimodal Learning   & +    & +       &     & +   \\ 
    \bottomrule
    \end{tabular}
    \label{tab_previous_work}
    \end{table}

\section{Methods}
    
    This section provides a detailed overview of the proposed Attention-Guided Concept Model (AGCM). We begin by detailing the selection and generation of multimodal concepts, a critical step before deploying any concept-based explainable model. Next, we focus on the visual modality, as it is the most widely used and complex modality, uniquely supporting explanations of what concepts contribute to predictions and where they are observed. Finally, we describe the multimodal architecture, addressing the challenges of multimodal alignment and co-learning. Using the audio-visual modality as an example, we demonstrate the framework's functionality and highlight its extendability to other signal-based modalities.
    
\subsection{Multimodal Concept Selection \& Generation} \label{sec_spa_concept}
    The selection of concepts or features plays a pivotal role in producing accurate and explainable results, whether in interpretable concept-based models or traditional feature-based models. In terms of explainability, the concept function - similar to features - acts as a key representation of the underlying data. Moreover, concepts must explicitly capture attributes that are highly relevant and meaningful to the task at hand. For example, in object detection, attributes such as color and shape are critical, while in bird classification, features like wing morphology or bill structure provide significant insights.
    
    Explaining spatial signals, such as those in the visual modality, involves two key aspects: spatial explanations and conceptual insights, which are particularly critical in explainable medical analysis \cite{barnett2021case} and affective XAI \cite{belharbi2024guided}. To address this, AGCM integrates spatial concepts, enabling the model to learn not only \textit{what} to focus on but also \textit{where} to focus.

    For the conceptual explanations (the \textit{what} question), key features such as facial muscle movements, gaze direction, and head pose are important for assessing and interpreting an individual's affective state \cite{adolphs2002recognizing, Bayliss2007Affective, xinyuFG24}. To address spatial explanations (the \textit{where} question), patch-level attention maps are trained alongside each concept in an end-to-end, fully supervised manner. This method allows the model to associate concept contributions with their exact spatial locations, thereby enhancing both interpretability and overall performance.

    Since manually annotated attention maps are not always available for large-scale datasets, the spatial maps are localized based on facial landmarks \cite{belharbi2024guided, ma2021landmark}. In this paper, we utilize an open-source landmark detector \cite{wang2020deep} for automatic landmark detection. According to the landmark locations, Regions of Interest (ROI) maps are generated for all AUs, which are subsequently used to supervise the spatial concept attention map throughout model training.

    To integrate these ROI maps into our transformer-based concept learning framework, they are transferred into patch-level representations, $\text{PatchMaps}[i]$, by performing average interpolation, as described in (\ref{eq_interpolation}). Here, $\text{AUMaps}[i](x_1, y_1)$ denotes the value of the $i$-th input map at position $(x_1, y_1)$. The terms $x'$ and $y'$ correspond to the patch indices in the $x$ and $y$ dimensions, while $S_x$ and $S_y$ denote the respective scaling factors.
    \begin{equation}\label{eq_interpolation}
        \text{PatchMaps}[i] = \frac{1}{S_x \cdot S_y} \sum_{x_1 = x' S_x}^{(x' + 1) S_x} \sum_{y_1 = y' S_y}^{(y' + 1) S_y} \text{AUMaps}[i](x_1, y_1)
    \end{equation}

\begin{figure}[t]
    \centering
    \includegraphics[width=0.99\columnwidth]{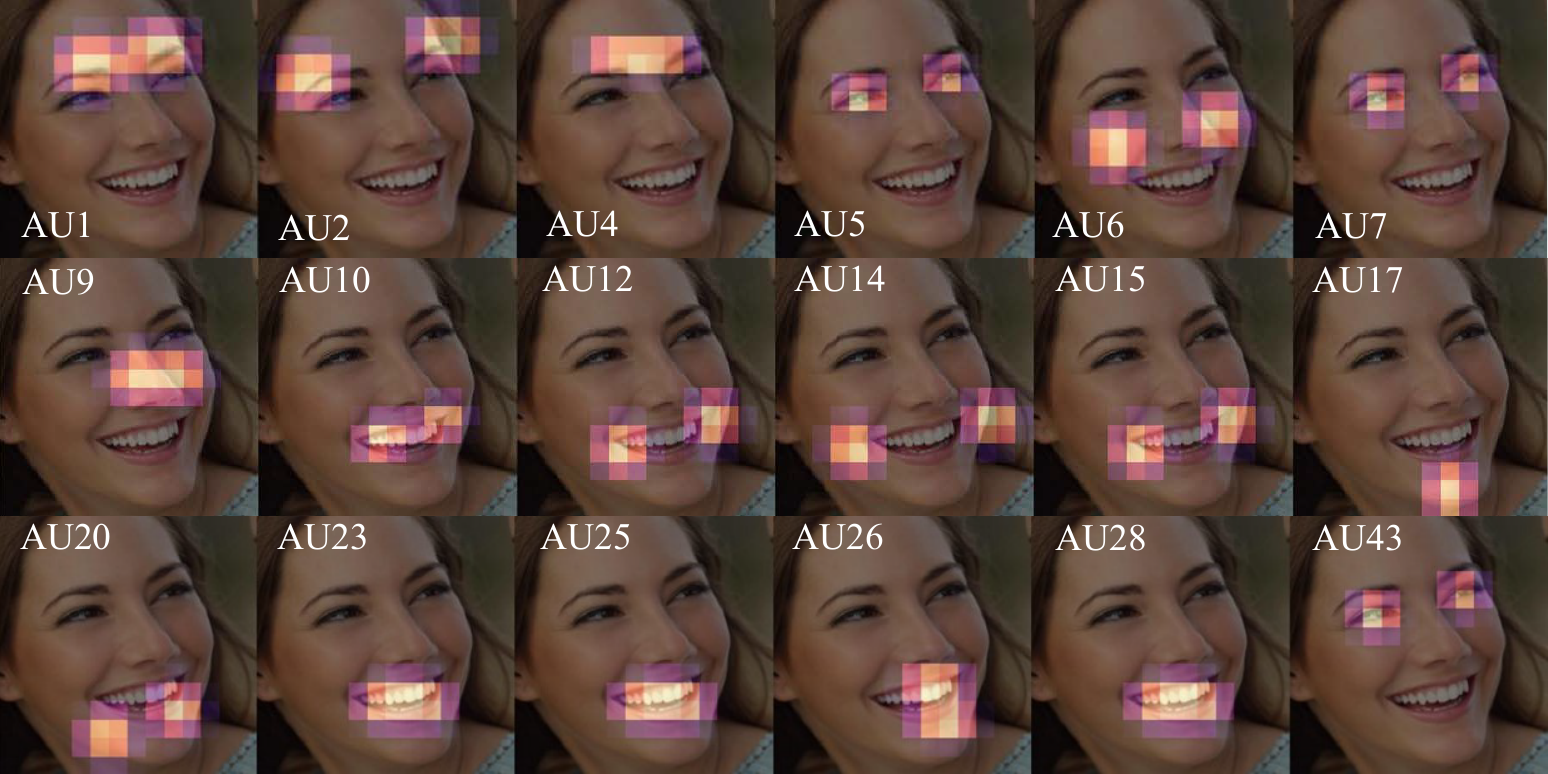}
       \caption{Example of patch-level AU map generated using landmark detection and average interpolation.}
    \label{fig_interpolation}
    \end{figure}

    Fig. \ref{fig_interpolation} presents an example of a patch-level AU map generated using landmark detection and average interpolation. In this map, patches with lighter colors indicate regions of higher importance, effectively highlighting the ROI for each AU. These maps are utilized as part of the ground truth to guide the model’s concept learning process via a concept map loss, ensuring the model's focus aligns with the actual spatial regions of interest during training.

    Other than spatial signals, temporal signals such as audio, Electrocardiogram (ECG), and Electroencephalogram (EEG) are often perceived as less complex in terms of dimensionality since they typically vary along a single axis (time). For these signals, stakeholders often prioritize conceptual insights (the \textit{what} question) over spatial interpretation. Temporal dependencies (the \textit{where} question in time) are naturally addressed by mechanisms like attention models or recurrence in sequential architectures, which excel at capturing temporal relationships.
    
    Using the widely used audio modality as an example, acoustic indicators such as pitch, loudness, and speech rate and their variations provide critical information by capturing subtle vocal variations that reflect emotional or cognitive states directly tied to the affective labels \cite{bachorowski1995vocal, Polzehl2011Anger, Yu2004Detecting, Adami2007Modeling, Grau1988The}. Providing conceptual insights into the decision-making process is essential for explaining predictions derived from these temporal signals.

\subsection{Visual Attention-Guided Concept Learning}

     %%%%%%%%%% Start SVG (AGCM) %%%%%%%%%%%%
    \begin{figure*}[th]
    \centering
    \includegraphics[width=1.8\columnwidth]{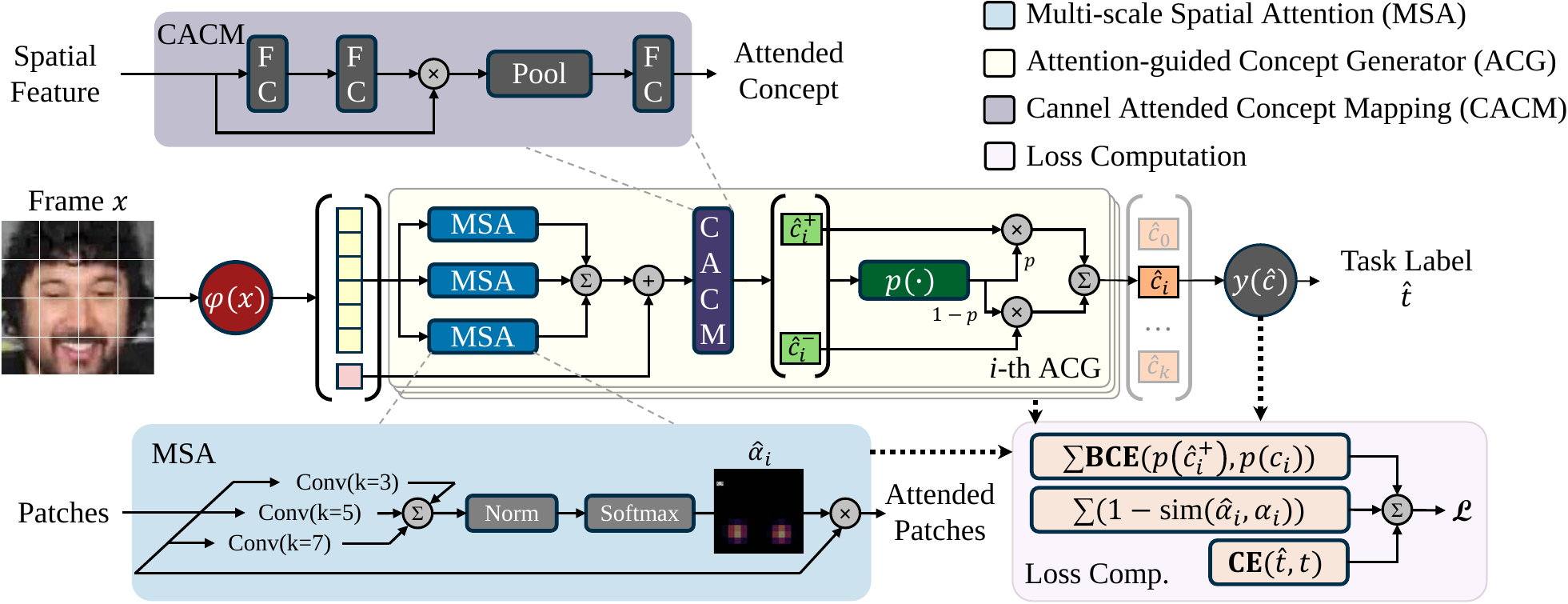}
       \caption{The architecture of our proposed Attention-Guided Concept Model (AGCM) for the spatial visual modality. The model uses a transformer backbone $\varphi(\cdot)$ to convert the facial image $x$ into a patch-level representation. The Attention-Guided Concept Generator (ACG) applies spatial-channel attention with a Multi-scale Spatial Attention (MSA) block and Channel Attended Concept Mapping (CACM), which together capture attention across both spatial and feature dimensions.
       The MSA block focuses on spatial features at multiple scales, enhancing the model's ability to capture both fine and coarse details. For instance, the concept of the cheek region may benefit from a larger attention area compared to the eye region. Three MSA heads are used to capture diverse spatial patterns within an image, each generating a concept attention map $\hat{a}_{i}$. These maps are weighted and summed to produce the final concept attention map, which is used to update the concept map loss during training.
       CACM further improves the model's focus on the most informative features along the channel dimension, ensuring robust feature selection across multiple channels.
       A concept probability generator $p(\cdot)$ computes the probability of each activated concept, facilitating concept supervision by showing the contribution of individual concepts to the predicted label. Notably, ACG considers both activated and inactivated concept embeddings, as the absence of certain concepts (e.g., AUs) can provide additional information about a subject's facial expression. The predicted activated concepts, $\hat{c}_{i}^{+}$, and inactivated concepts, $\hat{c}_{i}^{-}$, are weighted by their respective probabilities from $p(\cdot)$, then concatenated and passed to the one-layer fully-connected task predictor $y(\cdot)$ to generate the final task label $\hat{{\textit{t}}}$. 
       During loss computation, the model optimizes its performance using the task loss, concept probability loss, and concept map loss associated with the spatial concept attention, ensuring a strong explainability of the model's decision-making process giving not only \textit{what} key concepts contribute the most to the prediction but also \textit{where} these concepts appear. }
    \label{agcm_framwork}
    \end{figure*}
    %%%%%%%%%% END of SVG (CEM-based FER Framework) %%%%%%%%%%%%
  
    Given the complexity and the inherent differences between the spatial visual signal and other temporal signals, AGCM first focuses only on training the visual concept through attention-guided concept learning. This architecture leverages spatial concept supervision and concept attention to interpret the model's decision-making process by determining not only \textit{what} key concepts contribute the most to the prediction but also \textit{where} these concepts appear.

    As illustrated in Fig.~\ref{agcm_framwork}, the proposed Attention-Guided Concept Model (AGCM) is designed to enhance both the accuracy and explainability of the concept-based models. The model begins by processing the input facial image $x$ through a transformer backbone $\varphi(\cdot)$, which converts the image into a patch-level representation. This representation effectively captures local and global features by dividing the image into patches and is essential for subsequent processing.

    The core component of AGCM is the Attention-Guided Concept Generator (ACG), which integrates two attention mechanisms: Multi-scale Spatial Attention (MSA) and Channel Attended Concept Mapping (CACM). The MSA block focuses on spatial features at multiple scales, enabling the model to capture both fine-grained and coarse details within the image. For example, recognizing the concept of the cheek region may require a broader attention area compared to the eye region. To achieve this, three MSA heads are employed to capture diverse spatial patterns, each generating a concept attention map $\hat{a}_{i}$. These maps are then weighted and summed to produce a final concept attention map, which is utilized to update the concept map loss during training. 
    
    Complementing the spatial attention, CACM enhances the model's focus along the channel dimension. By applying attention to the most informative feature channels, CACM ensures robust feature selection across multiple channels, which is crucial for accurately interpreting complex facial expressions.

    The proposed framework also includes a concept probability generator $p(\cdot)$ that computes the probability of each activated concept. This mechanism facilitates concept supervision by quantifying the contribution of individual concepts to the predicted label. Importantly, ACG considers both activated and inactivated concept embeddings because the absence of certain concepts (e.g., deactivation of AUs) can also provide valuable information about one's facial expressions. The $i$-th predicted activated concepts, $\hat{c}_{i}^{+}$, and inactivated concepts, $\hat{c}_{i}^{-}$, are weighted by their respective probabilities from $p(\cdot)$. The probability score $p$ indicates the likelihood that the activated concept contributes to the final prediction. These are then concatenated and passed to the task predictor $y(\cdot)$, which is a one-layer fully connected network, to generate the final task label $\hat{\textbf{\textit{t}}}$. Therefore, it is designed to be adaptable and expandable to any discrete or continuous concepts, given that appropriate concept annotations are available.

    During loss computation, the model optimizes performance through a combination of losses: task loss, $\mathcal{L}_{t}$, concept probability loss, $\mathcal{L}_{c}$, and concept map loss, $\mathcal{L}_{m}$, associated with spatial concept attention. The task loss, $\mathcal{L}_{t}$, is computed using Cross Entropy (CE), while the concept probability loss, $\mathcal{L}_{c}$, is derived from the sum of Binary Cross Entropy (BCE) across all concepts. Instead of relying on Mean Square Error (MSE), the concept map loss, $\mathcal{L}_{m}$, uses Cosine Similarity (sim) to emphasize spatial pattern alignment rather than strict value matching. Therefore, the total loss, $\mathcal{L}$, is formulated as:
   \begin{equation}\label{eq_loss}
    \mathcal{L} = \text{CE}(\hat{t}, t) + \sum_{i=1}^{n} \text{BCE}(p(\hat{c}_i^+), c_i) + \sum_{i=1}^{n} (1 - \text{sim}(\hat{a}_i, a_i)).
    \end{equation}
    Here, $t$ is the ground truth task label, $c_i$ is the label of the $i$-th concept, and $a_i$ is the $i$-th concept attention map, while $n$ denotes the total number of used concepts.
    
    This comprehensive optimization strategy ensures that the AGCM framework achieves high accuracy while maintaining explainability in its predictions.

\subsection{Expandable Multimodal AGCM Concept Fusion}
    
     %%%%%%%%%% Start SVG (AGCM) %%%%%%%%%%%%
    \begin{figure*}[th]
    \centering
    \includegraphics[width=1.3\columnwidth]{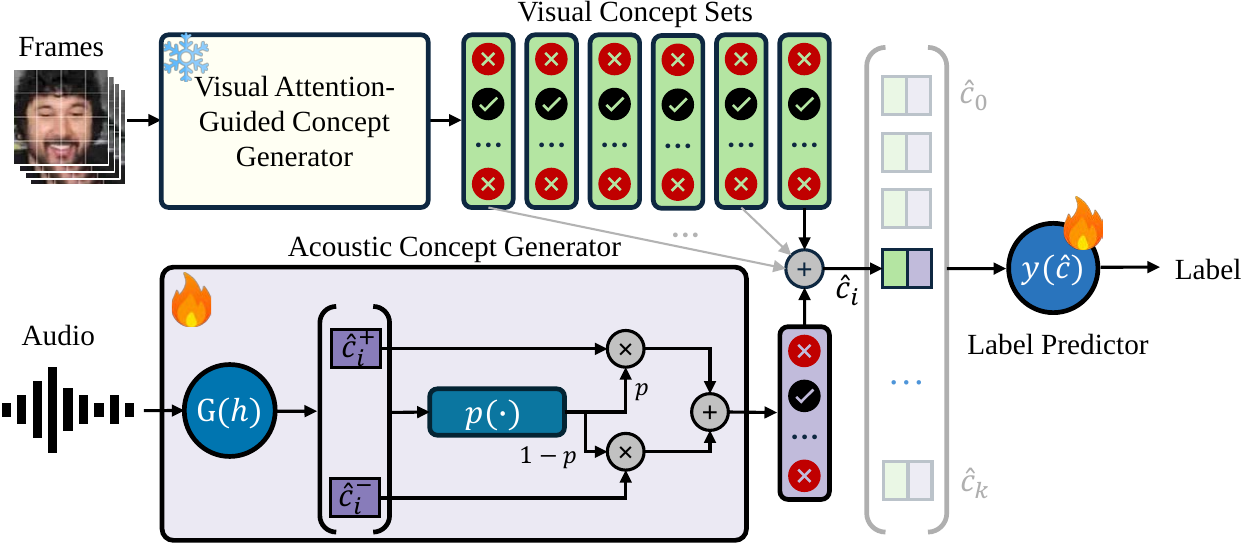}
       \caption{
       In the multimodal fusion stage, the pre-learned visual branch functions as a Visual Attention-Guided Concept Generator. The parameters of the Visual Attention-Guided Concept Generator are frozen to ensure reliable visual concept predictions. On the audio side, an Acoustic Concept Generator (ACG) processes the audio input, generating activated ($\hat{c}_{i}^{+}$) and inactivated ($\hat{c}_{i}^{-}$) acoustic concept embeddings via an acoustic feature extractor $G(\cdot)$. The probability of each concept's activation is computed using an acoustic concept probability generator $p(\cdot)$. The acoustic concept embeddings are concatenated with their corresponding visual concept set and passed through a sequential bottleneck layer {$\hat{c}_{0}$, ...$\hat{c}_{k}$}, where $k$ represents the number of samples in the sequence. For a given video clip, it is assumed that acoustic concepts are shared across all frames. A sequence-to-sequence label predictor $y(\cdot)$ is then used to capture the contextual relationships between frames to generate the final by-frame task label. Importantly, the AGCM framework is inherently extendable to other temporal modalities by adding additional branches to accommodate new data inputs, as long as the appropriate data and annotations are available.
       }
    \label{fusion_framwork}
    \end{figure*}
    %%%%%%%%%% END of SVG (CEM-based FER Framework) %%%%%%%%%%%%

    Alignment, fusion, and co-learning are three primary challenges in multimodal learning, involving the ability to identify, combine, and transfer knowledge across different modalities \cite{baltruvsaitis2018multimodal}, particularly in the context of interpretable AC \cite{cortinas2023toward}. After training the visual concept branch in the first stage, AGCM integrates visual information with any other temporal modalities through concept fusion. In this work, we demonstrate AGCM is an expandable multimodal architecture, using the most commonly used audio-visual fusion as an example, which involves identifying audio information using an acoustic concept generator and joining and transferring knowledge via a late fusion concept-label classifier. 

    As shown in Fig.~\ref{fusion_framwork}, the fusion stage builds upon the visual-based branch from the previous stage. During the fusion stage, the task predictor from the visual branch is removed, transforming it into a Visual Attention-Guided Concept Generator. This visual generator is responsible for extracting and predicting key visual concepts, including AUs, gaze direction, and head poses. To ensure stability and reliability in visual concept prediction, the parameters of the visual branch are frozen, preventing further modifications during the audio-visual training phase. This approach allows the model to harness pre-learned visual knowledge without overfitting, facilitating robust integrated learning across diverse input modalities.

    In parallel with the visual concept branch, the fusion stage introduces an audio brunch with an Acoustic Concept Generator (ACG) to process the audio input. This generator identifies relevant audio information using an acoustic feature extractor, denoted as $G(\cdot)$. These features are then mapped into activated ($\hat{c}_{i}^{+}$) and inactivated ($\hat{c}_{i}^{-}$) acoustic concept embeddings. The probability of activation for each concept is computed through an acoustic concept probability generator $p(\cdot)$, which quantifies the likelihood of each acoustic concept being present in the input. 

    For downstream applications, the audio branch can be replaced or expanded to incorporate other temporal modalities, such as Electrocardiogram (ECG), Electroencephalogram (EEG), or Electrodermal Activity (EDA), provided the appropriate data and annotations are available.

    Once the visual and temporal concepts are extracted, they are aligned and concatenated to form a unified multimodal representation. In this architecture, a key assumption is made: for a given video clip, temporal concepts are shared across all frames. This allows the model to maintain temporal coherence in the audio stream while aligning it with frame-specific visual features. The bottleneck layer serves to compress the multimodal information, ensuring that only the most relevant aspects of the fused representation are retained for further processing. The concatenated concepts are then passed through a sequential bottleneck layer, denoted as { $\hat{c}_{0}$, ..., $\hat{c}_{k}$ }, where $k$ represents the number of samples in the sequence. 

    To capture the temporal and contextual relationships between frames, the fusion branch employs a sequence-to-sequence concept-label predictor $y(\cdot)$, using a transformer architecture. This predictor is designed to handle sequential data, leveraging the temporal dependencies between consecutive frames in a video. By utilizing sequential learning, the model effectively integrates and co-learns multimodal information across time, improving the accuracy of by-frame predictions. This is particularly important for tasks where affective signals evolve over time, such as conversational engagement estimation or mental health assessment.

    The final task label is generated on a per-frame basis, with the model predicting the affective state for each frame in the video sequence. The combination of multimodal concept embeddings allows the VA-AGCM to provide robust and accurate predictions, as it captures a wider range of cues that contribute to affective behavior. Notably, the AGCM framework is readily extendable to other temporal modalities by incorporating additional branches for new data inputs.

\section{Experimental Evaluation and Results}

    Given the intricate nature and wide-ranging applications of AC tasks, we initially employed Facial Expression Recognition (FER) in both visual and audio-visual settings to validate the efficacy of our proposed AGCM framework, considering its well-established datasets and baseline models. We quantitatively evaluate the task and concept-level performance of AGCM on three large-scale FER datasets, and provide qualitative visualizations of the visual and multimodal conceptual explanations, demonstrating the framework’s robustness through occlusion experiments and an ablation study.

\subsection{Datasets}

    We employ three popular benchmark datasets, including RAF-DB and AffectNet with visual modality and Aff-Wild2 with audio-visual data. 

    \textbf{RAF-DB} \cite{li2017reliable} is a widely-used static FER dataset sourced from the internet, containing 6 basic emotion labels (Surprise, Disgust, Fear, Happiness, Sadness, Anger), and a Neutral label. The dataset includes 12,271 images in the training set and 3,068 images for testing.

    \textbf{AffectNet} \cite{mollahosseini2017affectnet} is one of the largest FER datasets, comprising 420,000 facial images annotated with categorical emotion labels. We utilize AffectNet-8, which consists of 291,651 manually labeled images with 8-class emotion labels (Neutral, Happy, Angry, Sad, Fear, Surprise, Disgust, and Contempt). In addition, we employ AffectNet-7, which contains 287,401 images annotated with seven emotion labels (excluding Contempt). The test set contains approximately 3,500 images.

    \textbf{Aff-Wild2} \cite{kollias2019expression} is a large-scale in-the-wild dataset specifically designed for FER and AU detection. It includes over 2.7 million frames from 564 videos with 554 subjects. We use the \textbf{by-frame} FER subset which is manually labeled with 8-class discrete emotions (Neutral, Anger, Disgust, Fear, Happiness, Sadness, Surprise, Other). It also provides manual annotation of 12 AUs. 

\subsection{Concept Generation Setup}
    AGCM is designed to be flexible and extendable to all kinds of discrete or continuous concepts, provided suitable concept annotations are available. In this work, we use the most commonly used audio-visual pair as an example. For the audio concepts, pitch, pitch variation, pitch stability (Jitter), loudness, loudness variation, and speech rate are used. For the visual modality, AUs, gaze direction, and head pose are used. 
    
    Unlike AUs, which are binary in nature (activated or inactivated), gaze, head pose and acoustic concepts are continuous and must be mapped into a probability space to fit the concept-based framework. Specifically, gaze concepts are defined as the degree of direct forward gaze in both horizontal and vertical planes, where $1$ represents directly looking forward and $0$ indicates looking elsewhere. Head pose concepts capture deviations in yaw (head shake) and pitch (head nod). These gaze and head pose concepts are scaled to the range $[0, 1]$ to fit within the concept probability generator, and corresponding heatmaps are generated based on facial landmarks, similar to Section \ref{sec_spa_concept}.

    All acoustic concept labels are normalized to the range $[0, 1]$ before AGCM training. To ensure alignment with the visual concepts, the video data is split into one-second clips (FPS=30), with a 33ms stride applied to capture temporal information effectively. 
    For clips containing complete silence, both pitch and loudness are set to $0$, indicating no contribution from the audio modality. Variations in loudness and pitch are calculated using their first-order derivatives, representing the rate of change for these acoustic features, while the Jitter is inherently a percentage. For videos featuring multiple speakers, the audio track for each subject will be individually separated to minimize noise and ensure clarity. 
    
    Furthermore, the AGCM framework is flexible and can incorporate other temporal modalities with continuous or discrete values, provided the appropriate data and annotations are available.

\subsection{Implementation Details}
     Our experimental setup is summarized as follows: AGCM utilized a pre-trained Vision Transformer as the backbone feature extractor \cite{dosovitskiy2020image}. Similar to \cite{xinyuFG24}, the backbone was pre-trained on VGGFace2 \cite{cao2018vggface2} for the facial recognition task. After pre-training, the classification header was removed and replaced with the AGCM workflow. Facial images were cropped from the video dataset using the InsightFace detector \cite{an2022killing}. To prevent overfitting, the preprocessing stage incorporated random data augmentation techniques, including horizontal flipping, random rotation, and random erasing.

     For datasets lacking AU annotations, we utilized OpenFace 2.0 \cite{baltrusaitis2018openface} to automatically extract 18 Action Units (AUs), which served as intermediary concepts in our proposed framework. All models were trained for 100 epochs, with early stopping to avoid overfitting, and optimized using the Adam optimizer (learning rate set to 0.0001). The AGCM generated concepts using a Dropout rate of 0.01 and Leaky-ReLU activation. The concept probability and map loss weights were set to $1$, ensuring a balanced focus on both conceptual explanation and task prediction. 

     AGCM used HuBERT \cite{hsu2021hubert} feature extractor for the audio input. During concept fusion, the learning of the vision branch was frozen, and the Acoustic Concept Generator (ACG) was fine-tuned for 100 epochs, with early stopping (learning rate set to 0.0001). All experiments were conducted on a workstation equipped with dual 48GB Nvidia RTX 6000 Ada GPUs, running a Linux-based PyTorch environment. For quantitative performance evaluation, we report the average performance over four random seeds. 

\subsection{Evaluating Visual-based AGCM}

    \begin{table*}[th]
    \centering
    \caption{Performance comparison of various models in terms of overall accuracy (\%) on RAF-DB, AffectNet-7, and AffectNet-8. The proposed AGCM framework consistently outperforms feature-based, map-based, and concept-based interpretable models. Notably, AGCM also surpasses state-of-the-art black-box models, offering superior performance without sacrificing conceptual interpretability.}
    \begin{tabular}{llllccc}
    \toprule
    Type                               & Model       & Year & Architecture                & RAF-DB         & AffectNet-7    & AffectNet-8    \\ \midrule
    \multirow{6}{*}{Black-box ML}      & AFR  \cite{savchenko2023adr}         & 2023 & EfficientNet                & 90.05          & 66.51          & 63.13          \\
                                       & CL-TransFER \cite{yang2024cl} & 2024 & Transformer                 & 91.33          & \textbf{67.86}          & \textbf{64.69}          \\ 
                                       & HAM \cite{tao2024hierarchical}        & 2024 & Attention                   & 91.92          & 66.97          & 63.82          \\
                                       & Poster++ \cite{mao2024posterpp}   & 2024 & Transformer                 & 92.21          & 67.49          & 63.77          \\
                                       & CEPrompt \cite{zhou2024ceprompt}   & 2024 & Transformer                 & 92.43          & 67.29          & 62.74          \\
                                       & S2D \cite{chen2024static}        & 2024 & Transformer                 & \textbf{92.57}          & 67.62          & 63.76          \\ \midrule
    Feature-based ML                   & FC          & 2024 & 3-layer FC                  & 67.04          & 40.23          & 37.11          \\ \midrule
    \multirow{3}{*}{Map-based XAI}     & TS-CAM \cite{gao2021ts}           & 2021 & Transformer + CAM & 86.70          & 62.28          & 58.99          \\
                                       & \multirow{2}{*}{Att-Map \cite{belharbi2024guided}}  & 2024 & CNN + Map Attention         & 88.88          & \textbf{62.45}          & \textbf{61.30}          \\
                                       &             & 2024 & Transformer + Map Attention & \textbf{91.03}          & 62.28          & 61.19          \\ \midrule
    \multirow{2}{*}{Concept-based XAI} & CEM \cite{xinyuFG24}        & 2024 & Concept Embedding           & 91.05          & 67.60          & 63.70          \\
                                       & AGCM        & 2024 & Spatial Attention Concept   & \textbf{94.40} & \textbf{69.45} & \textbf{65.62} \\ \bottomrule
    \label{tab_acc_rafdb}
    \end{tabular}
    \end{table*}

    Given the complexity and necessity of determining not only \textit{what} key concepts contribute the most to the prediction but also \textit{where} these concepts appear, we begin by evaluating the visual branch on RAF-DB and AffectNet. To assess the efficiency of the proposed AGCM framework against the previous feature-based and explainable models, we compared this work with a feature-based model, end-to-end map-based explainable models (with CNN and ViT backbones), previous concept-based explainable models, and the state-of-the-art black-box model without explicit model explainability. 
    
    The feature-based model uses only handcrafted features (e.g., AUs) as input, and a 3-layer Fully Connected (FC) neural network with ReLU activation, matching the complexity of AGCM's task predictor.

    Table \ref{tab_acc_rafdb} presents the overall accuracy of various models on RAF-DB, AffectNet-7, and AffectNet-8. The proposed AGCM framework achieves the highest accuracy across all datasets, with 94.40\% on RAF-DB, 69.45\% on AffectNet-7, and 65.62\% on AffectNet-8. These results demonstrate AGCM's significant improvement over the classic feature-based methods, particularly on RAF-DB (+27.36\%) and AffectNet-8 (+28.51\%). AGCM also outperforms state-of-the-art black-box transformer models including S2D \cite{chen2024static} and Poster++ \cite{mao2024posterpp}, providing gains of 1.83\% on RAF-DB and 1.86\% on AffectNet-8 compared to S2D. This highlights AGCM's ability to match and exceed black-box model performance while maintaining conceptual explainability. Furthermore, AGCM demonstrates superior results compared to interpretable map-based approaches, with a 3.37\% improvement on RAF-DB and over 4\% on AffectNet. When compared to the previous concept-based model \cite{xinyuFG24}, AGCM shows consistent gains across all datasets, benefiting from its spatial concept and attention learning.

    \begin{table}[t]
    \centering
    \caption{Class-wise performance comparison (\%) of the proposed AGCM and the transformer-based Poster++ \cite{mao2024posterpp} on RAF-DB and AffectNet-8. AGCM gives a more balanced performance along all classes, resulting in higher average accuracy. }
    \begin{tabular}{l|cc|cc}
    \hline
             & \multicolumn{2}{c|}{RAF-DB}     & \multicolumn{2}{c}{AffectNet-8} \\ \cline{2-5} 
             & AGCM           & POST++         & AGCM           & POST++         \\ \hline
    Anger    & \textbf{94.53} & 88.27          & \textbf{66.05} & 60.20          \\
    Disgust  & \textbf{82.43} & 71.88          & \textbf{61.58} & 58.00          \\
    Fear     & \textbf{87.50} & 68.92          & 63.00          & 63.00          \\
    Happy    & \textbf{97.47} & 97.22          & \textbf{79.42} & 76.40          \\ 
    Sad      & \textbf{93.51} & 92.89          & 65.01          & \textbf{66.80} \\
    Surprise & 89.51          & \textbf{90.58} & 62.99          & \textbf{65.60} \\
    Contempt & -              & -              & \textbf{64.08} & 59.52          \\
    Neutral  & \textbf{93.68} & 92.06          & \textbf{62.76} & 60.60          \\ \hline
    Avg.     & \textbf{91.23} & 85.97          & \textbf{65.61} & 63.77          \\ \hline
    \end{tabular}
    \label{tab_classwise_performance}
    \end{table}

    Table \ref{tab_classwise_performance} presents the class-wise performance comparison between the proposed AGCM framework and the black-box transformer-based Poster++ model \cite{mao2024posterpp} on RAF-DB and AffectNet-8. The results clearly demonstrate the effectiveness of AGCM in delivering a more balanced performance across all FER classes than Poster++, resulting in higher average accuracy on both datasets. The result shows the efficiency of considering conceptual prior knowledge, such as AUs and ROI maps, into the training process to quantify the individual concept's contribution towards predicting the label.

    On RAF-DB, AGCM consistently outperforms Poster++ across nearly all emotion classes, particularly in challenging categories such as Anger and Disgust, where AGCM achieves significant improvements of +6.26\% and +10.55\%, respectively. AGCM also demonstrates superior performance in the Fear class (+18.58\%), while maintaining competitive accuracy in easier classes like Happy and Neutral.
    
    Similarly, on AffectNet-8, AGCM provides improved accuracy in most categories, including notable gains in Anger (+5.85\%), Disgust (+3.58\%), and Happy (+3.02\%). Although Poster++ marginally outperforms AGCM in the Sad and Surprise categories, AGCM still delivers a more balanced overall performance, as evidenced by the higher average accuracy (+1.84\%).
    
    The consistent class-wise improvements offered by AGCM highlight its ability to maintain strong performance across both datasets, even in the presence of class imbalance and data variability. More importantly, AGCM not only surpasses Poster++ in terms of average accuracy but also achieves these gains while preserving the model's interpretability, which is essential for applications requiring both performance and transparency.

\subsection{Evaluating Multimodal AGCM}\label{sec_eval_multimodal}

    \begin{table}[t]
    \centering
    \caption{Performance comparison of various models in terms of average F-1 score (\%) on the uni- and multimodal Aff-Wild2 dataset.  }
    \begin{tabular}{lllll}
    \toprule
    Type                       & Model                & Arch.           & Data & F-1            \\ \midrule
    \multirow{5}{*}{Black-box} & DAN \cite{wen2023distract}                 & Attention       & V    & 40.10          \\
                               & AFR \cite{savchenko2023adr}                 & EfficientNet    & V    & 42.10          \\
                               & MAE \cite{ma2023unified}                 & MAE             & V    & 44.60          \\
                               & TCN \cite{zhou2023leveraging}                 & Transformer     & V/A  & 41.38          \\
                               & MMAE \cite{zhang2023multi}                & MAE+Transformer & V/A  & \textbf{48.93} \\ \midrule
    Feature                    & FC                   & FC              & V    & 25.27          \\ \midrule
    \multirow{3}{*}{Map}       & TS-CAM \cite{gao2021ts}              & Transformer     & V    & 37.05          \\
                               & \multirow{2}{*}{Att-Map \cite{belharbi2024guided}} & CNN             & V    & \textbf{41.92} \\
                               &                      & Transformer     & V    & 40.87          \\ \midrule
    \multirow{3}{*}{Concept}   & CEM \cite{xinyuFG24}                 &                 & V    & 42.60          \\
                               & AGCM                 &                 & V    & 44.95          \\
                               & AGCM                & Multimodal Fusion  & V/A  & \textbf{47.52} \\ \bottomrule
    \end{tabular}
    \label{tab_affwild2}
    \end{table}
    
    AGCM framework is designed to be expandable to multimodal inputs and concepts. In this work, we use the most commonly used audio-visual dataset as an example, demonstrating the AGCM's capacity for aligning and co-learning information from spatial and temporal modalities. 
    
    To evaluate the overall performance of the AGCM framework in a multimodal context, we conducted comprehensive experiments using the audio-visual Aff-Wild2 dataset.

    Table \ref{tab_affwild2} presents the performance comparison in terms of the average F-1 score on the Aff-Wild2 dataset. The proposed AGCM framework consistently outperforms feature-based, map-based, and concept-based interpretable models. Notably, AGCM in a multimodal setting achieves competitive results compared to state-of-the-art black-box models that leverage multimodal data, while maintaining conceptual explainability.

    Specifically, AGCM attains an F-1 score of 47.52\% by combining visual and audio inputs, outperforming visual-only AGCM (+2.57\%) and CEM (+4.92\%), showing that generally it works better in the multimodal setting. In comparison to feature-based models, AGCM demonstrates a significant improvement (+22.25\%), emphasizing the effectiveness of concept-level multimodal alignment and co-learning. While CNN-based map models \cite{belharbi2024guided} show stronger performance among map-based approaches, they still lag behind AGCM (by 3.03\%) and AGCM (by 5.6\%).
    
    The black-box MMAE model \cite{zhang2023multi} achieves the highest F-1 score of 48.93\%, largely due to its use of a pre-trained transformer (Masked Autoencoder or MAE), which is computationally expensive, time-consuming, and lacks interpretability. In contrast, the competitive results of AGCM highlight its ability to deliver robust performance while offering interpretability, which is a key advantage over black-box methods, even in the real-world multimodal context.

\subsection{Concept Efficiency}

    \begin{table}[t]
    \centering
    \caption{Concept Alignment Score (CAS) in percentage for all tasks. The score for no concept serves as a comparison. NOXI refers to the engagement estimation task in Section \ref{sec_noxi}.}
    \begin{tabular}{lcccc}
    \toprule
                & No Concept & CEM   & AGCM-V  & AGCM-AV \\ \midrule
    RAF-DB      & 66.10      & 78.62 & 82.36 & -     \\
    AffectNet-7 & 67.51      & 78.43 & 84.33 & -     \\
    AffectNet-8 & 66.29      & 78.01 & 83.52 & -     \\
    Aff-Wild2   & 65.09      & 77.36 & 81.29 & 81.46 \\
    NOXI        & 63.58      & 76.50 & 80.83 & 82.11 \\ \bottomrule
    \end{tabular}
    \label{tab_cas}
    \end{table}

    The efficiency of predicted concepts is a critical metric for both performance as well as explainability. To evaluate the reliability of learned concept representations, we employ the Concept Alignment Score (CAS) \cite{zarlenga2022concept}, which measures how well the predicted concepts align with their corresponding ground truth labels. Unlike traditional accuracy, which struggles with defining thresholds between ``activated'' and ``inactivated'' concepts, CAS uses homogeneity scores and clustering algorithms to assess the proximity of predicted concepts to ground truth, providing a more robust measure of concept alignment.

    As shown in Table \ref{tab_cas}, models without concept supervision (No Concept) serve as a baseline for comparison. The proposed framework in visual (AGCM-V) and audio-visual (AGCM-AV) contexts outperform the previous CEM models \cite{xinyuFG24}, which give higher CAS across all datasets, indicating their superior ability to learn meaningful and aligned concepts for both visual and audio modalities. 

\subsection{Human Interpretable Conceptual Explanation}
    
    In addition to achieving competitive performance compared to black-box deep learning models, a significant advantage of concept-based frameworks lies in their ability to offer clear, human-interpretable conceptual explanations grounded in domain-specific knowledge, making them accessible to even non-AI experts.

\subsubsection{Spatial Conceptual Explanation}
    Compared to the map-based approaches that only give one activation map as an explanation, AGCM combines the advantage of both concept-based and map-based models, which not only identifies \textit{where} the model focuses during inference but also explains \textit{what} specific facial behaviors the model is focusing on. 

\begin{figure}[t]
    \centering
    \includegraphics[width=0.99\columnwidth]{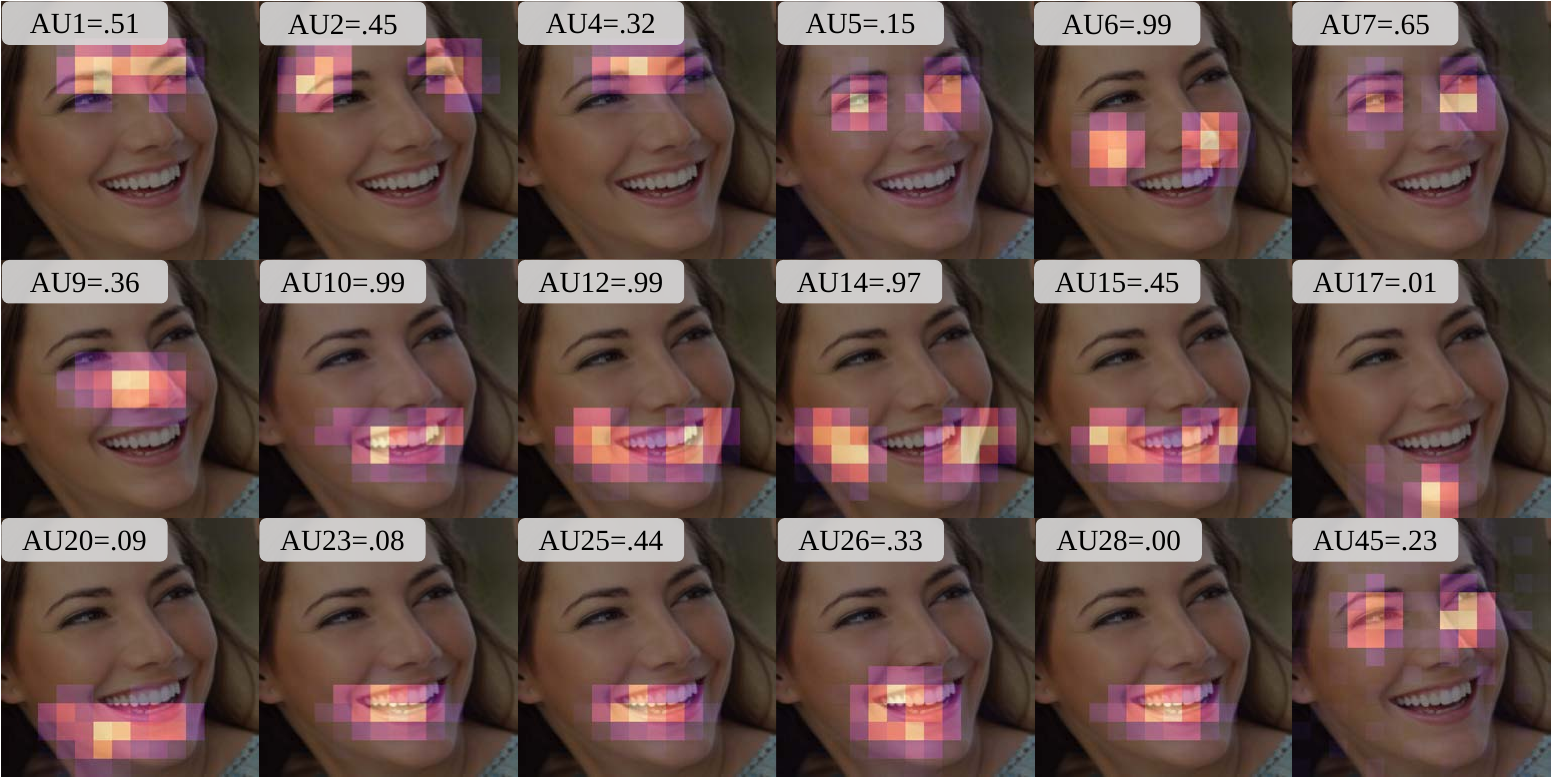}
       \caption{AGCM offers human-interpretable and intuitive explanations by presenting the contribution of each concept to the prediction alongside its spatial location. The numbers indicate the predicted probability scores for all concepts. During inference, the proposed AGCM generates attention maps for all concepts and assigns probability scores based on the highlighted regions. Concepts with higher probabilities (e.g., AU12) indicate greater contributions to the final label, while concepts with lower probabilities (e.g., AU28) are suppressed by AGCM's concept generator. }
    \label{fig_heatmap_pred}
    \end{figure}

    Fig. \ref{fig_heatmap_pred} illustrates the spatial concept explanations generated by the proposed AGCM for a facial image classified as ``Happiness'' from the AffectNet test set. During inference, AGCM produces attention maps for all relevant concepts and assigns probability scores based on the areas of the face highlighted in the maps. Concepts with higher probabilities, such as AU12 (Lip Corner Puller), are identified as making a significant contribution to the final classification, while those with lower probabilities, such as AU28 (Lip Suck), are effectively suppressed by the concept generator, reducing their influence on the predicted label. Compared to the map-based XAI that gives only a single attention map as the explanation, as in Fig. \ref{fig_intro}, the proposed model focuses on every possible expression indicator all over the facial region and then assigns the concept score to further indicate its contribution to a specific affective label, efficiently overcoming the trade-off between explainability and performance. 

    To simplify the visualization of the overall conceptual explanation, we proposed a weighted concept attention map $\bar{\alpha}$ that combines $i$-th predicted attention heatmaps $\hat{\alpha}_{i}$ with its corresponding concept probability $\hat{p}_{i}$, as given in (\ref{eq_norm_map}). Here, \textit{Norm} represents the min-max normalization, $n$ is the total number of concepts, and $\mathbb{I}(\hat{p}_{i} \geq \rho)$ is an indicator function that includes only concepts with probabilities exceeding the threshold $\rho$. We set $\rho = 0.5$ to visualize all activated concepts.
    \begin{equation}\label{eq_norm_map}
        \bar{\alpha} = \textit{Norm}\left(\sum_{i=1}^{n} \hat{\alpha}_{i}\cdot \hat{p}_{i} \cdot \mathbb{I}(\hat{p}_{i} \geq \rho)\right)
    \end{equation}
    \begin{figure}[t]
    \centering
    \includegraphics[width=0.99\columnwidth]{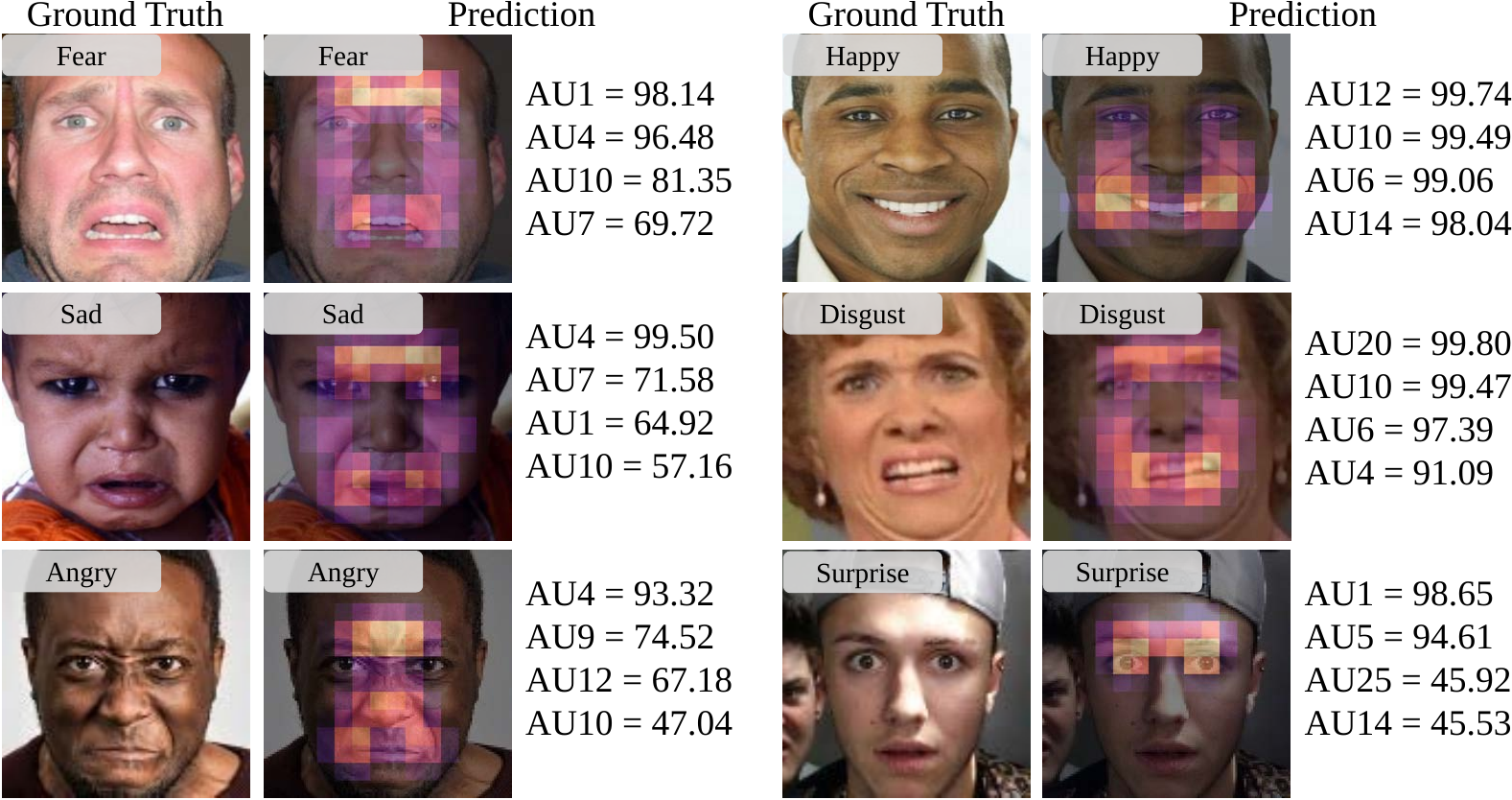}
       \caption{Example of the facial expression label prediction, top-4 concept probability predictions (\%), and weighted concept attention visualization from the AffectNet and RAF-DB test sets. The proposed AGCM framework offers intuitive interpretability by identifying the most contributing concepts to the prediction (addressing the \textit{what} question) and providing spatial explanations for where these concepts are observed (addressing the \textit{where} question).}
    \label{fig_example_correct}
    \end{figure}

    Fig. \ref{fig_example_correct} shows examples randomly selected from the AffectNet and RAF-DB test sets, illustrating the prediction of emotion labels alongside the top-4 concept probabilities (\%) and corresponding weighted concept attention visualizations. The AGCM framework accurately predicts class labels and provides insightful conceptual explanations through activated concept probabilities and attention heatmaps.

    In the ``Happy'' example, AU6 (Cheek Raiser), AU12 (Lip Corner Puller), and AU14 (Dimpler) are all strong indicators of happiness. AGCM efficiently focuses on the relevant facial areas while highlighting the contributions of these concepts. For the ``Anger'' expression, the model emphasizes AU4 (Brow Lowerer) and AU9 (Nose Wrinkler), which are the primary contributors to this emotion, with the attention maps focusing meaningfully on the brow and nose regions. These examples demonstrate AGCM's ability to combine robust performance with clear, human-interpretable conceptual explanations, making it readily applicable to domain-specific expertise.

\subsubsection{Spatial-temporal Conceptual Explanation}

    Another key advantage of the AGCM framework over previous interpretable approaches \cite{xinyuFG24, gao2021ts, belharbi2024guided}, is its ability to provide multimodal conceptual explanations from spatial and temporal data sources. Using audio-visual fusion as an example, AGCM enables the model to co-learn the information from multimodal data inputs, offering robust performance and better interpretable explanations in real-world multimodal contexts.

\begin{figure}[t]
    \centering
    \includegraphics[width=0.99\columnwidth]{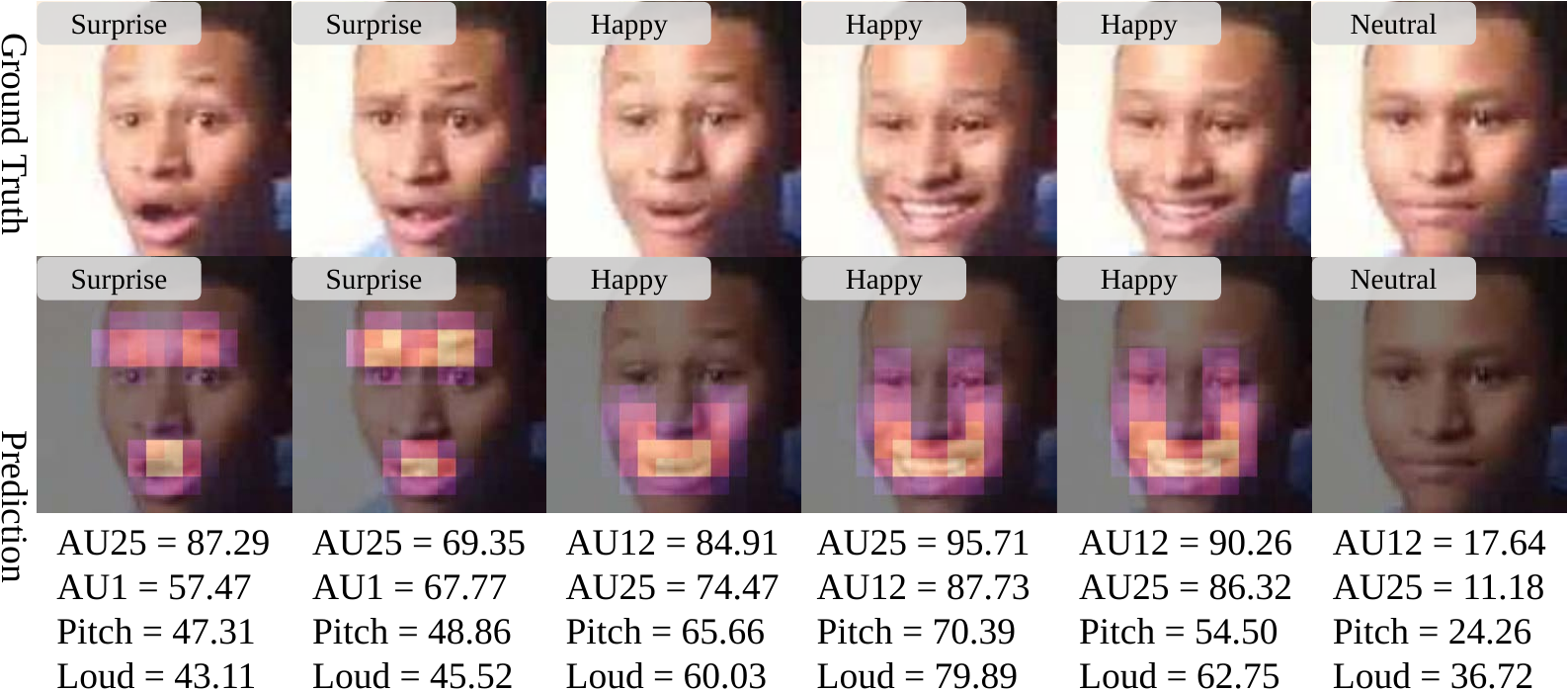}
       \caption{AGCM facilitates both the co-learning and interpretation of multimodal inputs. In addition to providing \textit{what}-\textit{where} explanations for the visual modality, AGCM offers \textit{what} the key conceptual insights into temporal signals. Temporal dependencies (\textit{where} in time) are naturally addressed through attention-based sequential learning. This figure shows an example from the Aff-Wild2 test set (~10 seconds), demonstrating this capability by including facial expression label predictions, top-2 AU probability predictions (\%), acoustic concept intensities (\%), and weighted concept attention visualizations. AGCM accurately predicts emotion transitions and downregulation while delivering human-interpretable conceptual explanations for both visual and acoustic modalities.}
    \label{fig_example_affwild2}
    \end{figure}

    Fig. \ref{fig_example_affwild2} illustrates an example of FER prediction on the Aff-Wild2 test set. We randomly selected this video clip to show approximately 10 seconds of data, which contains an emotional transition and downregulation event. Initially, the subject is in a ``Surprise'' state, where AGCM accurately identifies key visual concepts, such as AU25 (Lips Part) and AU1 (Inner Brow Raiser), which strongly indicate this emotion. 

    As the emotional transition occurs, AU1 decreases while AU12 (Lip Corner Puller) becomes dominant, signaling a shift toward a ``Happy'' state. Additionally, the model detects high intensities in pitch and loudness concepts, which are often associated with happiness because they reflect a sudden increase in physiological arousal, and are a natural reaction to pleasant and positive emotions \cite{kamilouglu2020good}. Toward the end of the clip, all concepts gradually decline, reflecting the downregulation of a high-intensity emotion back to a neutral state. AGCM enables the co-learning and interpretation of multimodal inputs by providing \textit{what}-\textit{where} explanations for the visual modality and identifying \textit{what} key conceptual insights derived from temporal signals. Additionally, temporal dependencies (\textit{where} in time) are handled through attention-based sequential learning during multimodal fusion, ensuring comprehensive interpretability across modalities.

\subsection{Robustness of the Explanation}
    
    To further evaluate the robustness of the model's explanations, we stress-test AGCM to explore its ability to handle challenging scenarios. Facial occlusion is a common challenge in real-world affective signal processing applications, particularly in in-the-wild datasets, where the subjects may wear VR glasses, causing upper-face occlusion, or masks, leading to lower-face occlusion. These occlusions present difficulties for affective computing, especially when providing conceptual or map-based explanations. The proposed AGCM framework addresses this limitation by generating weighted concept attention maps, which improve both the performance and the interpretability.
    
    To simulate real-world occlusion scenarios, we selected images from the Aff-Wild2 test set and manually occluded either the upper or lower face regions, re-evaluating the performance of the well-trained AGCM framework.

\begin{figure}[t]
    \centering
    \includegraphics[width=0.99\columnwidth]{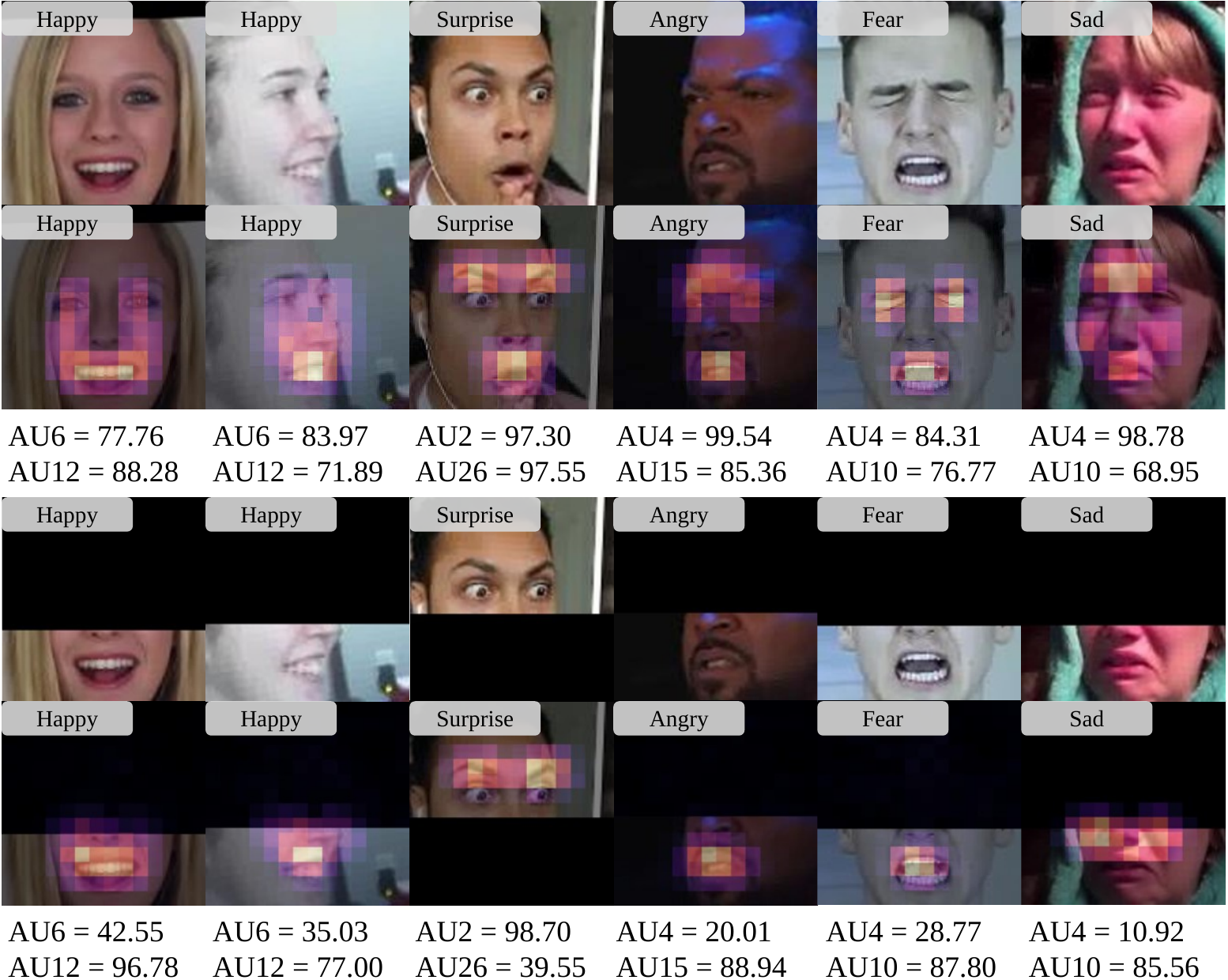}
       \caption{Example of the face occlusion samples, with the expression label prediction, representative concept probability predictions (\%) from the upper and lower-face region, and weighted concept attention visualization from the Aff-Wild2 test sets. After occlusion, AGCM adapts by shifting attention to the non-occluded areas, ensuring reliable predictions based on the remaining visible concepts. }
    \label{fig_concept_occ}
    \end{figure}

    As shown in Fig. \ref{fig_concept_occ}, we randomly selected samples with varying facial expressions, lighting conditions, and angles, then removed either the upper or lower face regions. Using the same well-trained AGCM model, we re-evaluated the predicted emotion labels, representative top concept probabilities, and the corresponding weighted concept attention maps. After occlusion, AGCM still accurately predicts the emotion by focusing on the unobstructed facial regions. In the ``Happy'' examples, the model shifts attention away from AU6 (Cheek Raiser), which is occluded and focuses more on AU12 (Lip Corner Puller), resulting in a correct prediction despite the occlusion. Similarly, in the ``Surprise'' example, AGCM downweights the contribution of the occluded AU26 (Jaw Drop) and instead focuses on AU2 (Outer Brow Raiser), another strong indicator of surprise. These results demonstrate AGCM’s robustness in handling occluded facial images while maintaining accurate and interpretable predictions.

    Hand-over-face occlusion presents an even more complex challenge than occlusion caused by glasses and masks, as the hand can often be misinterpreted as part of the face during model inference because one's hands often share similar textures with the face. To evaluate AGCM’s performance in such scenarios, we selected additional samples from the Aff-Wild2 dataset, which contains instances of hand-over-face occlusion.

\begin{figure}[t]
    \centering
    \includegraphics[width=0.99\columnwidth]{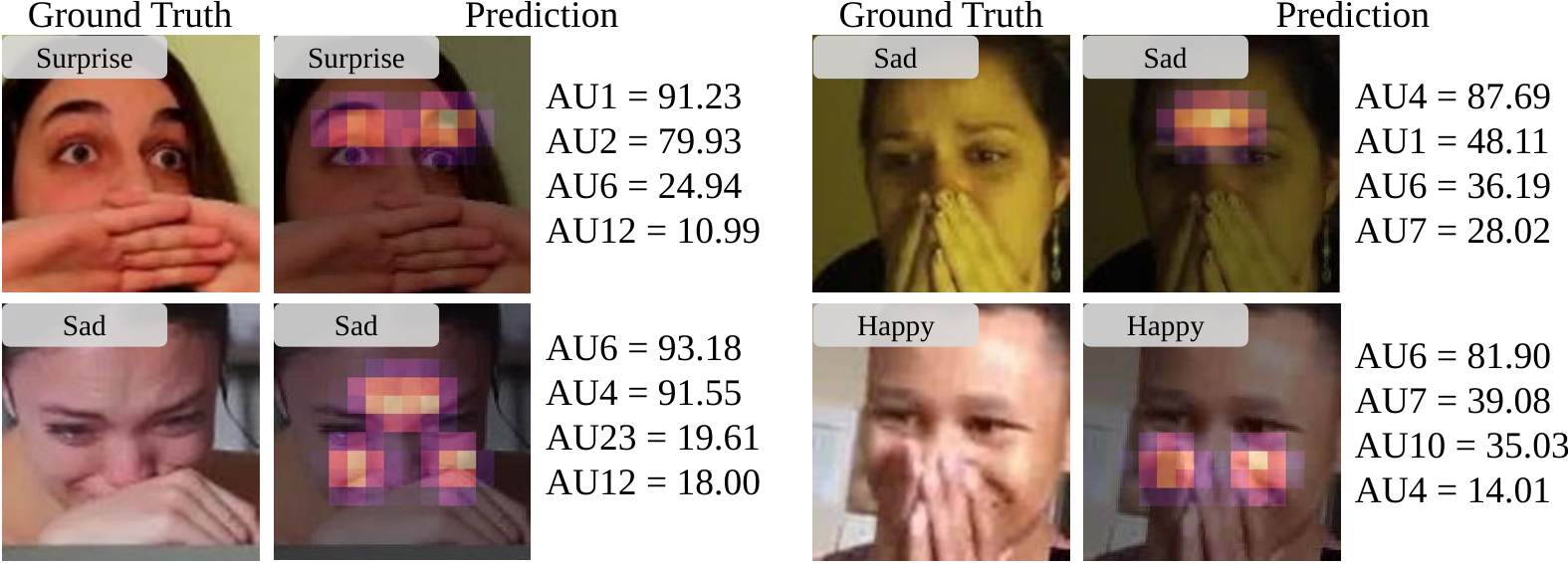}
       \caption{Example of hand-over-face occlusion, with the predicted facial expression label, top-4 concept probability predictions (\%), and weighted concept attention visualization. AGCM accurately focuses on the non-occluded regions and predicts the task label based on the available concepts, demonstrating its robustness in handling facial expressions with hand-over-face occlusion.}
    \label{fig_example_occ}
    \end{figure}

    Fig. \ref{fig_example_occ} shows test images featuring hand-over-face occlusion. Despite these occlusions, AGCM generates accurate emotion predictions by leveraging a few key concepts. For instance, in the ``Surprise'' example, even though the lower-face concepts are occluded, the model identifies high probabilities for upper-face indicators AU1 (Inner Brow Raiser) and AU2, leading to a correct prediction. Similarly, in the ``Happy'' example, AU6 (Cheek Raiser) alone is sufficient for the model to make this accurate prediction. 
    
    These stress-testing results demonstrate that AGCM effectively handles partial face occlusion and hand-over-face occlusion by focusing on unobstructed regions and leveraging spatial concept learning to emphasize visible concepts during training. This capability highlights AGCM's robust, concept-aware spatial explanations, enabling reliable predictions even in challenging scenarios.

\subsection{Ablation Study}
    Compared to the previous concept-based approaches, the proposed AGCM framework introduces four main components, including Multi-scale Spatial Attention (MSA), Multi-head Attention (MHA), Cannel Attended Concept Mapping (CACM), and Concept Map Loss (CML). As the evaluation of multimodal concept fusion has been given in Section \ref{sec_eval_multimodal}, this section provides an ablation study to show the efficiency of the visual-based AGCM framework. 

    \begin{table}[t]
    \centering
    \caption{Ablation study of the visual-based AGCM framework on RAF-DB and AffectNet-8 test set.}
    \begin{tabular}{cccc|cc}
    \toprule
    MSA & MHA & CACM & CML & RAF-DB   & AffectNet-8 \\ \midrule
    -   & -   & -    & -   & 90.47 & 62.58     \\ 
    +   & -   & -    & -   & 92.84 & 62.99     \\
    +   & +   & -    & -   & 93.26 & 63.10     \\
    +   & +   & +    & -   & 93.31 & 63.46     \\
    +   & +   & +    & +   & \textbf{94.40} & \textbf{65.62}     \\
    \bottomrule
    \end{tabular}
    \label{tab_abl}
    \end{table}

    Table \ref{tab_abl} presents the ablation study for the visual-based AGCM framework on RAF-DB and AffectNet-8. The baseline model without any components achieves 90.47\% on RAF-DB and 62.58\% on AffectNet-8. Adding Multi-scale Spatial Attention (MSA) improves performance significantly, reaching 92.84\% and 62.99\%. Introducing Multi-head Attention (MHA) further boosts accuracy to 93.26\% and 63.10\%, while Channel Attended Concept Mapping (CACM) provides a slight improvement to 93.31\% and 63.46\%. Finally, the full AGCM with Concept Map Loss (CML) achieves the best results, 94.40\% on RAF-DB and 65.62\% on AffectNet-8, demonstrating the cumulative benefit of these components in enhancing accuracy while maintaining explainability.

\section{AGCM for Interpretable Engagement Estimation}\label{sec_noxi}

    The generalizability of the framework to downstream applications is essential for establishing a trustworthy AC system. Real-life affective signal processing is inherently more ambiguous, complex, and diverse compared to the well-defined FER task. One good example is human-human interactions, where the conversational engagement score is designed to measure the level and rate of engagement between participants, illustrating the broader and more nuanced requirements of real-world AC applications.
    
    In this section, we use the NOvice eXpert Interaction (NOXI) dataset, a large-scale,  well-annotated human-human interaction dataset with the engagement label, to illustrate AGCM's generalization capacity in real-world AC contexts. We conduct both qualitative and quantitative evaluations, demonstrating that AGCM achieves robust performance by automatically identifying key indicators and highlighting essential concepts.

    NOXI \cite{cafaro2017noxi} is designed for the analysis of human interaction in real-world, cross-cultural settings. It includes video recordings of novice-expert interactions in eight languages (English, French, German, Spanish, Indonesian, Arabic, Dutch, and Italian), with AU capture via Microsoft \textit{Kinect} \cite{zhang2012microsoft}. The dataset spans over 50 hours of video and is annotated with a \textbf{by-frame} engagement score ranging from 0 to 1. For our experiments, we utilize 76 videos (over 1.5 million frames) for training and 20 videos (over 500,000 frames) for testing.

\subsection{Generalizing AGCM for Engagement Estimation}

    \begin{table}[t]
    \centering
    \caption{Performance comparison of various models in terms of Concordance Correlation Coefficient (CCC) on the uni- and multimodal Noxi dataset.}
    \begin{tabular}{lllcc}
    \toprule
    Type                       & Model   & Arch.          & Data & CCC           \\ \midrule
    \multirow{3}{*}{Black-box} & TCA \cite{he2024tca}    & Attention      & V/A   & 0.73          \\
                               & DCTM \cite{tu2023dctm}   & Transformer    & V/A    & 0.77          \\
                               & S2S \cite{yu2023sliding}     & Transformer    & V/A    & \textbf{0.83} \\ \midrule
    Feature                    & FC      & FC             & V    & 0.23          \\ \midrule
    \multirow{2}{*}{Map}       & TS-CAM \cite{gao2021ts} & Transformer    & V    & 0.36          \\
                               & Att-Map \cite{belharbi2024guided} &                & -    & -             \\ \hline
    Concept                    & CEM     &                & V    & 0.48          \\
                               & AGCM    &                & V    & 0.59          \\
                               & AGCM    & Concept Fusion & V/A  & \textbf{0.80} \\
    \bottomrule
    \end{tabular}
    \label{tab_noxi}
    \end{table}

    AGCM is highly generalizable to downstream AC applications by simply adjusting the configuration of the final task predictor. For example, in FER tasks, a classification header is utilized, whereas in continuous signal prediction tasks, a regression header is employed. This flexibility allows AGCM to adapt a wide range of affective computing applications.

    Table \ref{tab_noxi} shows the performance comparison of various models in terms of the Concordance Correlation Coefficient (CCC) on the Noxi dataset for continuous engagement estimation. CCC is used to evaluate continuous tasks by measuring the agreement between predicted and true values, accounting for both correlation and accuracy, making it ideal for engagement estimation tasks. The proposed multimodal AGCM framework with audio-visual concept fusion again outperforms feature-based and previous concept-based models, showing its outstanding state-of-the-art performance in downstream real-world engagement estimation tasks.

    In the unimodal setting, AGCM with visual concepts achieves a CCC score of 0.59, marking a substantial improvement over the unimodal CEM (+0.11) and feature-based model (+0.23). This result highlights the advantages of spatial concept learning while underscoring the limitations of feature-based models in addressing the complexities of affective signal processing.

    In the multimodal context, AGCM attains a performance of 0.80, demonstrating the significant benefits of co-learning multimodal knowledge. This is particularly valuable in complex real-world AC applications, such as engagement estimation, where multiple modalities are essential for capturing and understanding nuanced human behavior.

    Although the black-box S2S model \cite{yu2023sliding} slightly outperforms AGCM with a CCC of 0.03, AGCM underscores its ability to approximate state-of-the-art results while maintaining interpretability. The attention map-based models \cite{belharbi2024guided} are not well-suited to this task, as they rely on predefined mappings between AUs and labels, which are not available for continuous engagement estimation. Additionally, TS-CAM \cite{gao2021ts}, which is restricted to the visual modality, also performs poorly in engagement estimation.

    Meanwhile, the Concept Alignment Score (CAS), as shown in Table \ref{tab_cas}, illustrates that the AGCM framework with audio-visual co-learning not only maintains competitive performance compared to state-of-the-art black-box deep learning models but also delivers accurate conceptual explanations. 
    
    Therefore, this sophisticated interpretable framework maintains competitive performance without compromise. By simply configuring the AGCM classifier, the performance evaluation on engagement estimation demonstrates the strong generalizability of AGCM to a wide range of downstream applications beyond FER, making it both powerful and accessible for diverse affective computing tasks.

\subsection{AGCM Explainability in Engagement Estimation}
    \begin{figure}[t]
    \centering
    \includegraphics[width=0.99\columnwidth]{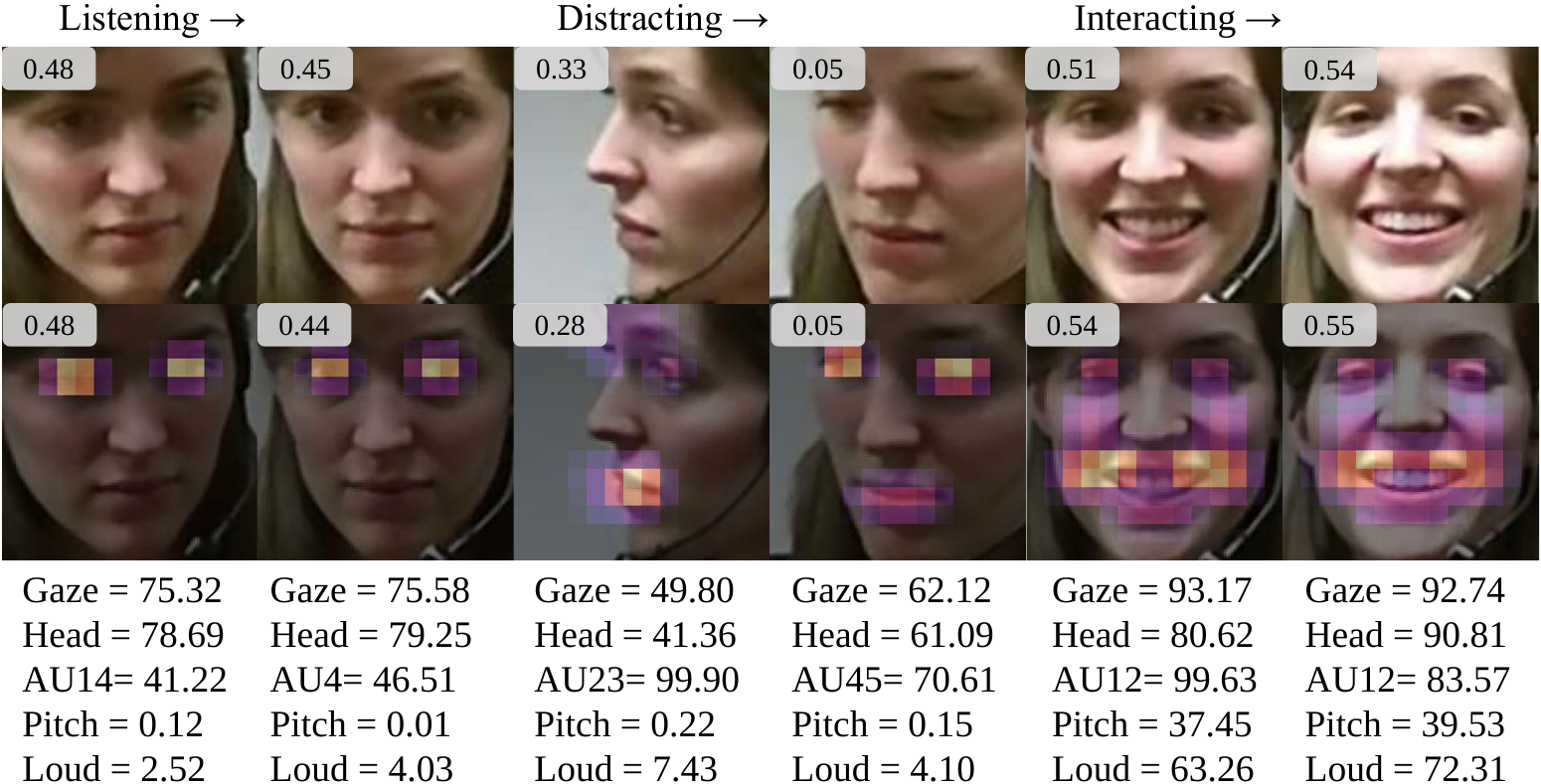}
       \caption{Example of the engagement estimation, gaze, head pose direction (mean degree of forward gaze or facing forward in x and y directions), top-1 AU probability predictions (\%), acoustic concept intensities (\%), and weighted concept attention visualization of a Noxi test sample (around 60 seconds). The proposed AGCM framework accurately predicts engagement transition for different states during conversation and provides meaningful visual and acoustic conceptual explanations.}
    \label{fig_example_noxi}
    \end{figure}

    Explainability becomes even more crucial in downstream AC applications compared to FER, given the inherent complexity of human behavior. In tasks such as engagement estimation, delivering domain-specific explanations is vital for non-AI stakeholders to understand and interpret the decision-making process.

\subsubsection{Explaining Engagement Transitions}

    To show AGCM's explanation and prediction capabilities in human-human engagement estimation, Fig. \ref{fig_example_noxi} presents an example from the NOXI dataset. This sample, randomly selected to cover approximately 60 seconds of data, highlights engagement transitions between listening, distraction, and interaction.

    At the beginning of the sequence, the subject actively listens with a direct gaze toward the speaker, as indicated by the high intensity of the conceptual direct gaze. In conversation-based engagement estimation, gaze direction and head pose are critical concepts for predicting and explaining engagement scores. Since the acoustic input is not prominent during the listening phase, the intensities of acoustic concepts remain low, which is expected as the audio track of each subject is recorded separately in this dataset.

    When distraction occurs, the subject shifts attention to a phone call or another person, causing the intensities of direct gaze and forward head pose concepts to decrease, which in turn lowers the engagement score. When the subject looks down with eyes only partly open, as evidenced by the activation of AU45 (Blink), the predicted engagement score reaches its lowest point, signifying that the subject’s attention is fully disengaged from the conversation.

    As the distraction ends and the subject re-engages with the speaker, positive facial expressions, such as AU12, become prominent, associated with increased engagement during interactions \cite{Greipl2021Facial, Savchenko2022Classifying}. The intensities of gaze and head pose concepts increase, and acoustic concepts begin to register, indicating the subject's regained focus and positive emotional feedback. This highlights the subject’s re-engagement in the conversation. 6
    
    AGCM effectively leverages these multimodal concepts to capture subtle changes in engagement states and ensure robust conceptual explainability during inference, demonstrating its strong generalizability to downstream AC applications with complex behavioral labels, extending beyond the scope of FER.

\subsubsection{Explaining Complex Human Behaviours}
    
    Real-world behavioral states are more complex than facial expressions. Higher-level affective states that share similar feature representations usually introduce ambiguity in task predictions, especially in feature-based models. 
    
    The proposed AGCM framework has an inherent advantage in differentiating nuanced states by autonomously learning and capturing both the concept-aware similarities and distinctions between these affective states. Take an example of human-human interaction, real-world applications often involve complex affective states that are more ambiguous and abstract compared to discrete emotions, such as distraction and cognitive load \cite{Krasich2018GazeBased}. 
    
\begin{figure}[t]
    \centering
    \includegraphics[width=0.95\columnwidth]{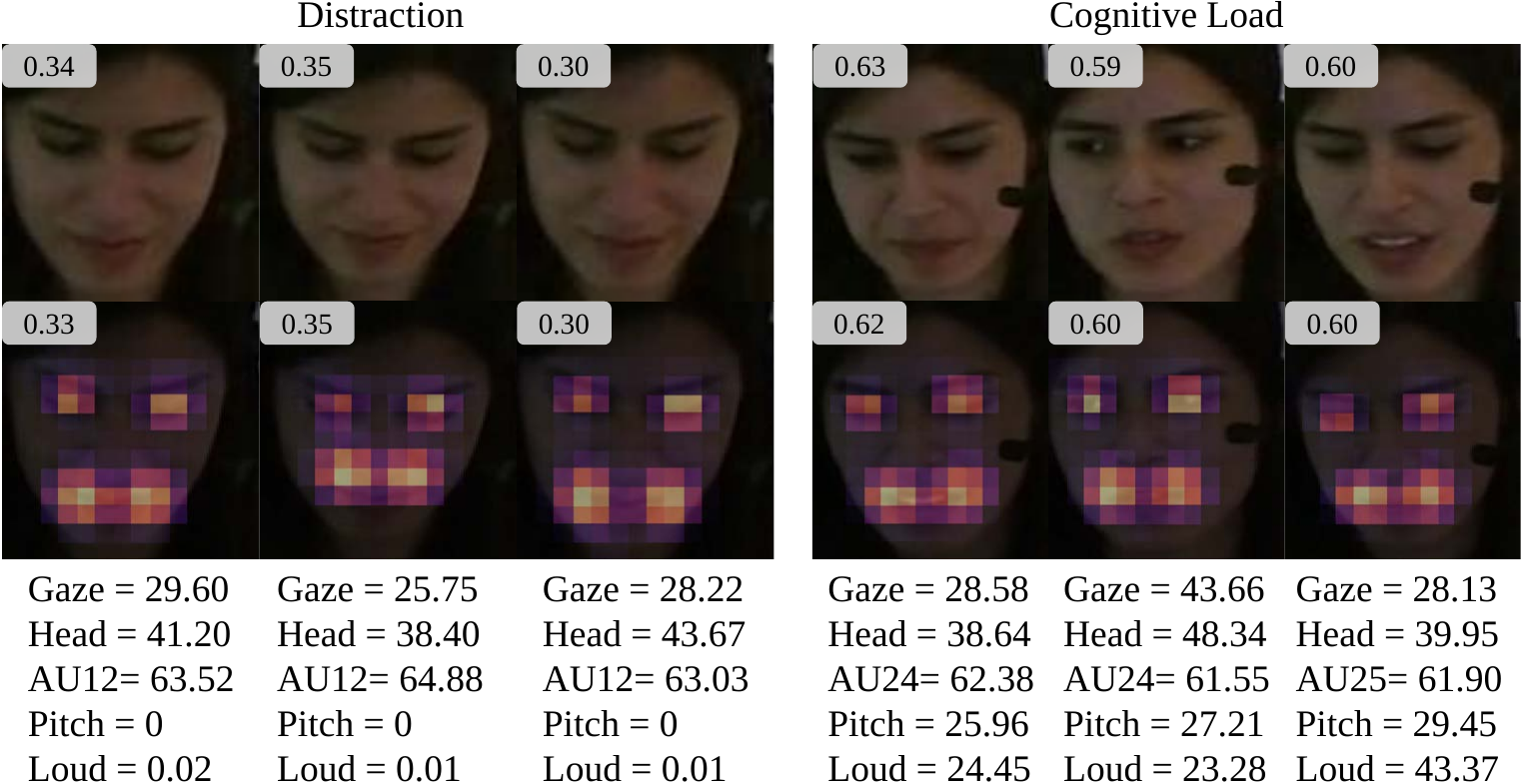}
       \caption{Example of the engagement estimation for distraction and cognitive load, with the prediction of gaze, head pose direction (mean degree of forward gaze or facing forward in x and y directions), top-1 AU probability predictions (\%), acoustic concept intensities (\%), and weighted concept attention visualization from Noxi dataset. AGCM differentiates between distraction and cognitive load according to efficient concept learning. }
    \label{fig_cognitive_load}
    \end{figure}

    Fig. \ref{fig_cognitive_load} provides an example of engagement estimation in the presence of distraction and cognitive Load. Distraction occurs when the subject's gaze drifts away, indicating mental disengagement. Conversely, cognitive load happens when the subject looks away while remaining engaged in processing information.  In feature-based affective computing models, these complex behaviors which share similar feature representations, can introduce ambiguity in task predictions.

    Thus, the proposed AGCM framework provides robust learning and explainability, even in complex behavioral states such as distraction and cognitive load. This demonstrates its effectiveness in capturing nuanced affective states, providing enhanced generalizability to complex downstream AC applications that are difficult to tackle using conventional methods.

\section{Conclusion \& Future Work}

    In this paper, we introduce the Attention-Guided Concept Model (AGCM), a multimodal concept-based interpretable framework that provides conceptual explanations of \textit{what} concepts contribute to the predictions and \textit{where} they are observed. AGCM is highly extendable to various spatial-temporal modalities, effectively addressing the challenges of multimodal alignment, fusion, and co-learning. The framework demonstrates strong generalizability and flexibility, making it well-suited for diverse real-world AC applications.
    
    We first validate the model's effectiveness in achieving both high performance and robust explanation through qualitative and quantitative evaluations on well-established FER datasets. Then, we demonstrate the generalizability of the AGCM framework to other complex real-world AC applications by extensive experiments on the human-human interaction task. We believe that AGCM establishes a foundation for creating future interpretable systems in downstream AC applications, such as psychology, psychiatry, digital behavior, and Human-Computer Interaction, with competitive performance and human-interpretable explanation.

    AGCM leverages the strengths of both feature-based models and deep black-box models to offer interpretable, high-performance predictions. However, explainability in affective computing remains an evolving area of research. We posit that model explanations should be tailored to end-users, such as psychologists and cognitive scientists. Therefore, we plan to incorporate a human-in-the-loop approach for affective XAI to further enhance model usability. Additionally, while AGCM is trained on large datasets, exploring XAI fairness in terms of gender, cultural, and age biases presents an interesting avenue for further investigation. In this paper, we assess various forms of occlusion using the Aff-Wild2 dataset; future improvements could be achieved by fine-tuning AGCM on occlusion-specific datasets to better handle such challenges. Generating text-based explanations via Large Language Models (LLM) may also give users extra insights. However, given the inherent complexity of LLMs, it is imperative to employ appropriate knowledge distillation techniques, particularly for cross-disciplinary stakeholders.

% {\small
% \bibliographystyle{ieee}
% \bibliography{egbib}
% }

{\small

 % must use main.bbl as the name, following the main.tex file
}
\vfill

\clearpage
% \onecolumn
% \appendix
\section*{Appendix}

\subsection{Expanded Discussion of AGCM}

    %%%%%%%%%% Start SVG (AGCM) %%%%%%%%%%%%
    \begin{figure*}[th]
    \centering
    \includegraphics[width=1.99\columnwidth]{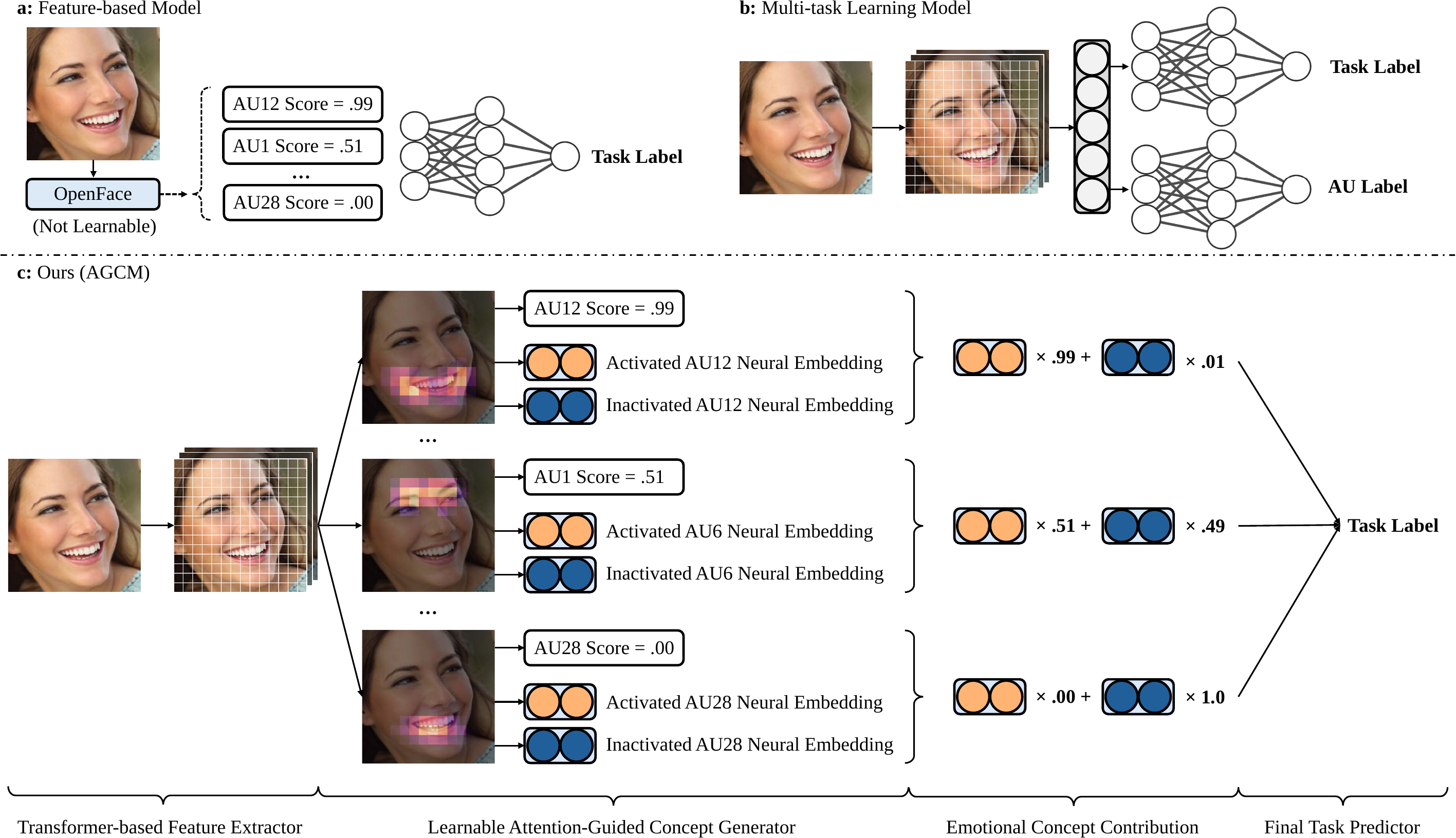}
       \caption{
       (a) Feature-based models rely on manual feature preprocessing using external automatic toolkits, such as OpenFace, which operate outside the model's training loop and are not learnable. These models map preprocessed features to task labels, risking the loss of valuable raw data information that could contribute to more comprehensive predictions.
       (b) Multi-task learning models train multiple tasks independently, with the learning of specific emotional tasks and AUs being uncorrelated and disconnected. As a result, AU predictions in multi-task learning cannot effectively explain the emotional predictions, limiting the interpretability of the model.
       (c): The proposed AGCM framework operates as follows: after feature extraction, the Attention-Guided Concept Generator creates learnable neural representations for both activated and inactivated concepts, along with their respective activation scores. It then computes the emotional concept contribution by combining the activated and inactivated embeddings for each concept. Parameter optimization for concept learning is conducted concurrently with task-label learning in an end-to-end manner, enabling the model to capture emotional concept contributions while effectively overcoming the trade-off between explainability and performance. }
    \label{fig_app_1}
    \end{figure*}
    %%%%%%%%%% END of SVG (CEM-based FER Framework) %%%%%%%%%%%%

    The use of handcrafted features, such as AU detections, has been ongoing for decades. These approaches mainly focus on automatically mapping the facial representation to a single numerical value, without fully accounting for the complexity of one’s affective state. Like in most of the feature-based approaches, relying solely on these numerical values for intricate AC tasks risks overlooking other emotion-related information conveyed by the subject, potentially degrading performance. Similarly, in multi-task learning—for instance, simultaneously predicting AU and expression—each classification head optimizes independently, rather than fostering mutually beneficial learning that emphasizes the relevance of AUs to facial expressions. 

    In contrast, as illustrated in Fig. \ref{fig_app_1}, the proposed AGCM framework enhances both model explainability and performance by bridging this gap. It employs an end-to-end learning strategy that quantifies the contributions of underlying emotion-related indicators to the final task prediction. By design, AGCM naturally advances traditional feature-based and multi-task AC approaches, where feature representations are either static or insufficient as explanations for task predictions.

\subsection{Embedding Size Ablation Study}

    %%%%%%%%%% Start SVG (AGCM) %%%%%%%%%%%%
    \begin{figure}[th]
    \centering
    \includegraphics[width=1\columnwidth]{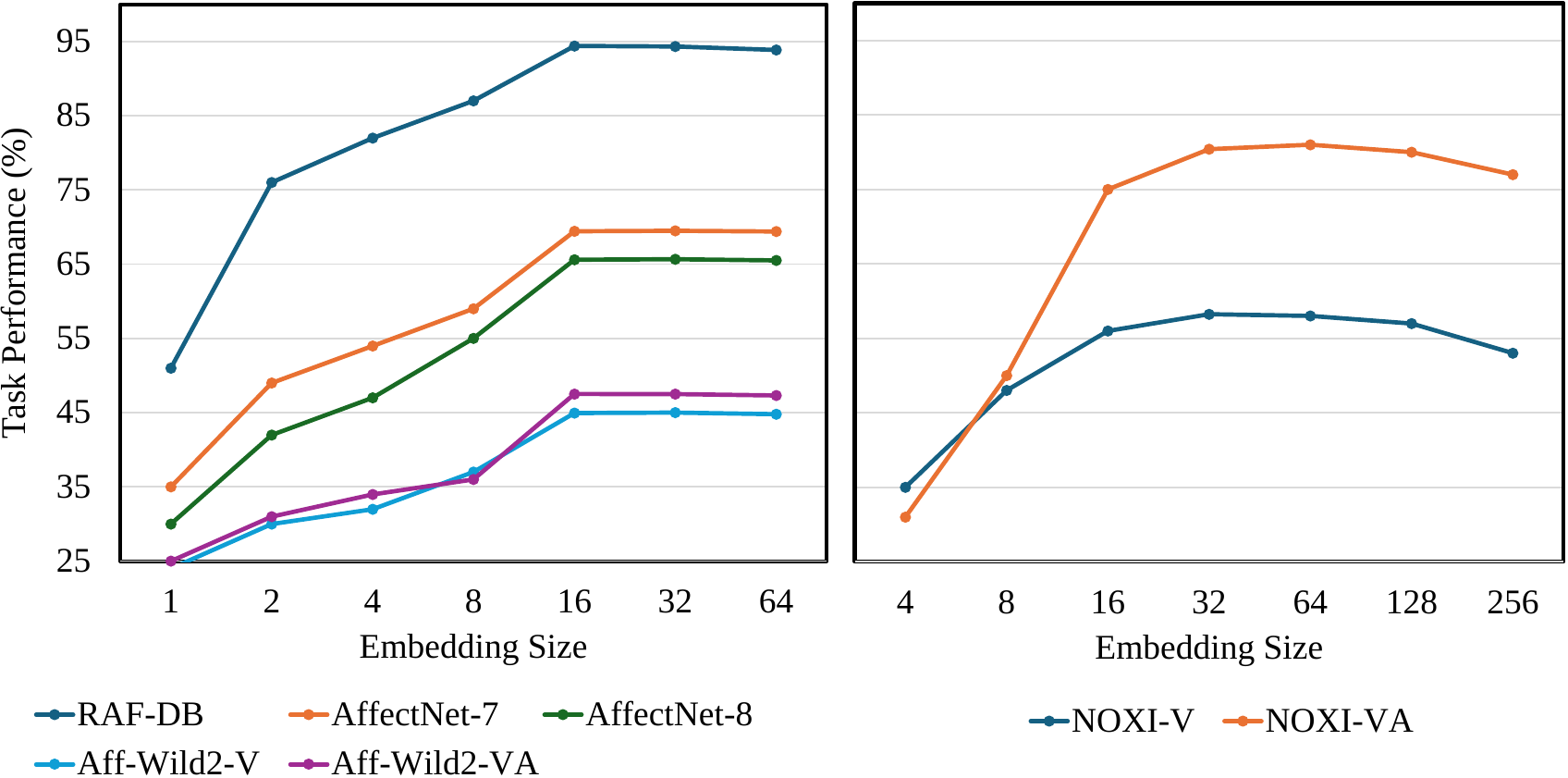}
       \caption{
       Task performance evaluation (\%) with different embedding sizes. For RAF-DB and AffectNet, the overall accuracy is reported. For Aff-Wild2 and NOXI, the F-1 score and CCC score are reported.}
    \label{fig_emb_size}
    \end{figure}
    %%%%%%%%%% END of SVG (CEM-based FER Framework) %%%%%%%%%%%%
    
    Previous studies have demonstrated that embedding size can impact the task performance of concept-based frameworks \cite{zarlenga2022concept}. The optimal concept size may vary depending on the task. In this work, we use an embedding size of 16 for all FER tasks and 32 for engagement estimation tasks. 

    Fig. \ref{fig_emb_size} shows the task performance across various embedding sizes. For both applications, performance initially improves with increasing embedding size. However, once the embedding size reaches the limitation of the model's learning capacity, further increases do not yield performance gains. Instead, larger embeddings may significantly raise the number of parameters, which can pose challenges for model training and deployment.

\subsection{Comparing End-to-end and By-step AGCM}
    
    To further assess the efficiency of the AGCM framework, we compare the end-to-end and by-step training strategies. In by-step AGCM, the model first optimizes a mapping function from the raw input to all intermediate concept scores. If the concepts include only AUs, this phase operates similarly to an AU detector, generating activation probabilities for all AUs. These AU probabilities are then combined with the embeddings in a subsequent optimization step to predict the final facial expression label separately.
    
    In by-step AGCM, the neural embeddings of intermediate concepts are not trainable during task learning. The parameter optimization treats the concept and task loss separately. This approach contrasts with end-to-end training, where a unified push-pull joint loss is employed to enhance both concept explainability and task performance simultaneously.

    \begin{table}[t]
    \centering
    \caption{Performance comparison (\%) of the end-to-end and by-step AGCM framework. For RAF-DB and AffectNet, the overall accuracy is reported. For Aff-Wild2 and NOXI, the F-1 score and CCC score are reported. }
    \begin{tabular}{cccc}
    \toprule
                & Data & End-to-end AGCM & By-step AGCM \\ \midrule
    RAF-DB      & V    & \textbf{94.40}      & 89.71   \\
    AffectNet-7 & V    & \textbf{69.45}      & 64.08   \\
    AffectNet-8 & V    & \textbf{65.62}      & 61.36   \\
    Aff-Wild2   & V    & \textbf{44.95}      & 39.10   \\
    Aff-Wild2   & V/A  & \textbf{47.52}      & 39.23   \\
    NOXI        & V    & \textbf{59.24}      & 52.01   \\
    NOXI        & V/A  & \textbf{80.39}      & 67.88   \\
    \bottomrule
    \end{tabular}
    \label{tab_by_step}
    \end{table}

    Table \ref{tab_by_step} presents a performance comparison between the end-to-end and by-step AGCM training strategies. Compared to the end-to-end approach, the by-step training strategy results in performance degradation across all datasets, with particularly notable declines in the multimodal AGCM framework, where separately learning concepts can lead to significant information loss from the raw data. Thus, we posit that jointly learning the concept and task label enhances both model explainability and task performance by compelling the model to explicitly supervise human-understandable features derived from domain-specific prior knowledge.

\subsection{Expanded Discussion of AGCM and Map-based XAI}

    Map-based XAI was originally designed for general ML tasks like object localization, where attention heatmaps serve as effective tools to indicate object locations \cite{gao2021ts}. In affective signal processing, however, spatial concept explanations offer significant advantages over map-based XAI by providing domain-specific insights alongside task performance improvements. Simply presenting an attention heatmap over a face region offers minimal value for domain experts in AC applications. For instance, two opposing indicators, AU12 (Lip Corner Puller) and AU15 (Lip Corner Depressor), appear in the same region of the face, making it insufficient to rely solely on attention maps for emotion interpretation. Instead, conceptual explanations that explicitly indicate the activation and contribution of specific AUs provide a more natural and informative approach to AC tasks.

    Recent map-based FER work \cite{belharbi2024guided} uses pre-generated AU maps based on emotion labels to guide model learning, depending on a strict mapping between AUs and facial expressions. For example, for images labeled as ``happiness,'' this approach restricts the model’s focus strictly to the AU6 and AU12 regions, regardless of whether these specific AUs are activated, ignoring other facial information that may contribute to the expression. This rigid mapping not only degrades performance but also proves limiting in downstream AC applications, such as engagement estimation or mental health assessment, where there is no clear mapping between AUs and affective labels.

\begin{figure*}[t]
    \centering
    \includegraphics[width=1.99\columnwidth]{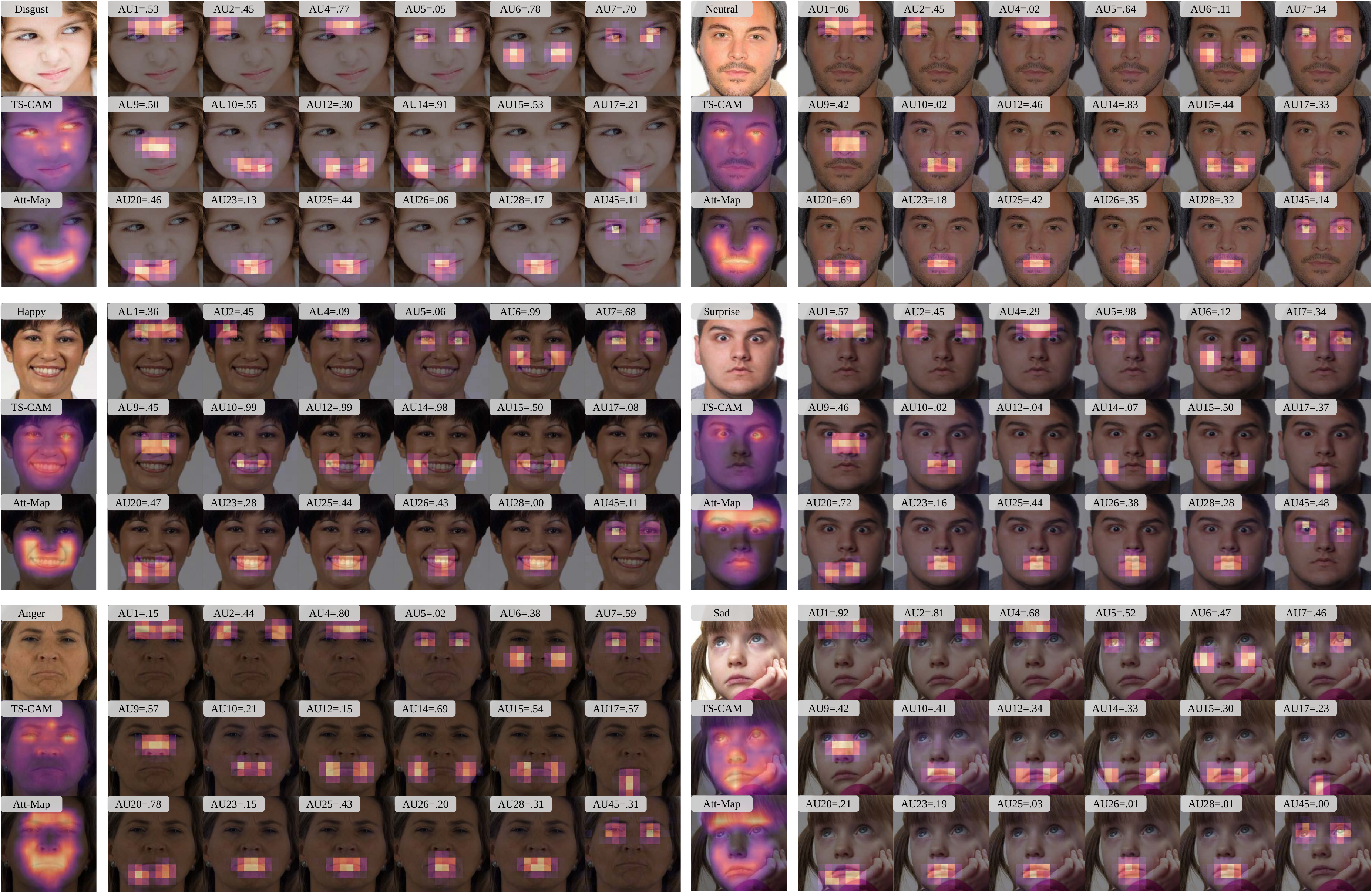}
       \caption{Explanation examples of map-based TS-CAM \cite{gao2021ts}, attention map-based FER (Att-Map) \cite{belharbi2024guided}, and the proposed AGCM framework. In addition to all concept locations, AGCM explicitly provides the contribution score of each concept, offering domain-specific insight into the model decision-making process. The images are randomly picked from the AffectNet test set. 
       }
    \label{fig_big_example}
    \end{figure*}

    Fig. \ref{fig_big_example} compares explanations provided by our proposed AGCM with those from two map-based XAI methods \cite{gao2021ts, belharbi2024guided}. The attention heatmaps from the map-based XAI approaches appear similar across different expression labels, offering insufficient interpretability for high-stakes AC applications. In contrast, AGCM not only localizes each AU but also quantifies its contribution to the final prediction, delivering richer insights into model predictions while achieving state-of-the-art task performance.

    \clearpage
    %%%%%%%%%% Start SVG (AGCM) %%%%%%%%%%%%
    %%%%%%%%%% END of SVG (CEM-based FER Framework) %%%%%%%%%%%%
\end{document}